\documentclass{article}
\usepackage{iclr2027_conference,times}
\usepackage{puredocbench_preprint}
\usepackage[utf8]{inputenc}
\usepackage[T1]{fontenc}
\usepackage{url}
\usepackage{booktabs}
\usepackage{longtable}
\usepackage{amsfonts}
\usepackage{nicefrac}
\usepackage{microtype}
\usepackage{xcolor}
\usepackage{colortbl}
\usepackage{array}
\usepackage{pifont}
\usepackage{tcolorbox}
\usepackage{fontawesome5}
\usepackage{tikz}
\usepackage{graphicx}
\usepackage{pdflscape}
\usepackage{enumitem}
\usepackage{multirow}
\usepackage{makecell}
\usepackage{wrapfig}
\usepackage{float}
\usepackage{placeins}
\usepackage{needspace}
\usepackage{hyperref}
\DeclareMathSizes{8.5}{8}{6}{5}
\hypersetup{colorlinks=true,citecolor=blue,linkcolor=blue,urlcolor=blue,pdfauthor={Zhiheng Li; Zongyang Ma; Jiaxian Chen; Jianing Zhang; Zhaolong Su; Yutong Zhang; Zhiyin Yu; Ruiqi Liu; Xiaolei Lv; Bo Li; Jun Gao; Ziqi Zhang; Chunfeng Yuan; Bing Li; Weiming Hu},pdftitle={How Far Is Document Parsing from Solved? PureDocBench: A Source-Traceable Benchmark across Clean, Degraded, and Real-World Settings}}
\newcolumntype{L}[1]{>{\raggedright\arraybackslash}p{#1}}

\definecolor{modelLink}{HTML}{087CA4}
\definecolor{cleanColumn}{HTML}{F3F7FC}
\definecolor{digitalColumn}{HTML}{FFF6EF}
\definecolor{realColumn}{HTML}{F2F8F1}
\definecolor{averageColumn}{HTML}{DDF3FA}
\definecolor{pipelineBand}{HTML}{C6DBF0}
\definecolor{e2eBand}{HTML}{E0CFEF}
\definecolor{vlmBand}{HTML}{C6E4DA}
\definecolor{resultStripe}{HTML}{F3F6F9}
\definecolor{cleanInk}{HTML}{285D8F}
\definecolor{digitalInk}{HTML}{A15A28}
\definecolor{realInk}{HTML}{247058}
\expandafter\def\csname modelname@dotsmocr\endcsname{DotsMOCR}
\expandafter\def\csname modelurl@dotsmocr\endcsname{https://huggingface.co/dots-studio/dots.mocr}
\expandafter\def\csname modelicon@dotsmocr\endcsname{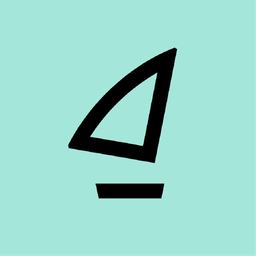}
\expandafter\def\csname modelname@dolphin-v2\endcsname{Dolphin-v2}
\expandafter\def\csname modelurl@dolphin-v2\endcsname{https://huggingface.co/ByteDance/Dolphin-v2}
\expandafter\def\csname modelicon@dolphin-v2\endcsname{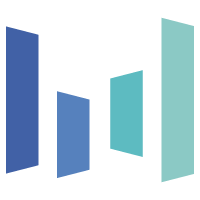}
\expandafter\def\csname modelname@monkeyocr-pro-3b\endcsname{MonkeyOCR-pro-3B}
\expandafter\def\csname modelurl@monkeyocr-pro-3b\endcsname{https://huggingface.co/echo840/MonkeyOCR-pro-3B}
\expandafter\def\csname modelicon@monkeyocr-pro-3b\endcsname{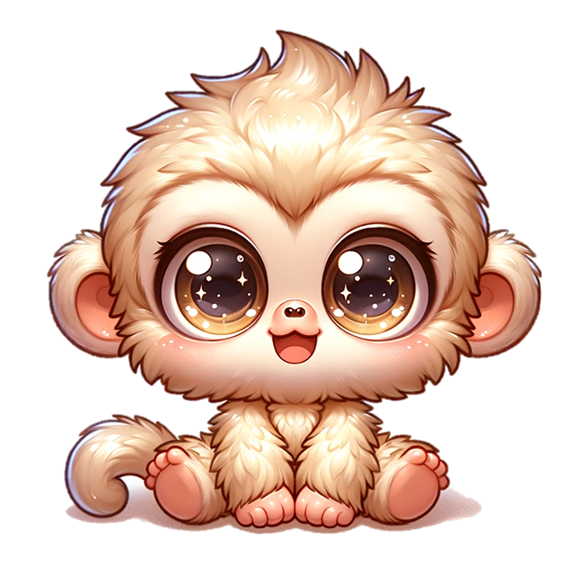}
\expandafter\def\csname modelname@youtu-parsing\endcsname{YouTu-Parsing}
\expandafter\def\csname modelurl@youtu-parsing\endcsname{https://huggingface.co/tencent/Youtu-Parsing}
\expandafter\def\csname modelicon@youtu-parsing\endcsname{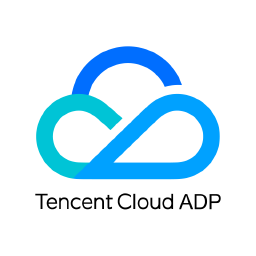}
\expandafter\def\csname modelname@mineru2-5-pro\endcsname{MinerU2.5-Pro}
\expandafter\def\csname modelurl@mineru2-5-pro\endcsname{https://huggingface.co/opendatalab/MinerU2.5-Pro-2604-1.2B}
\expandafter\def\csname modelicon@mineru2-5-pro\endcsname{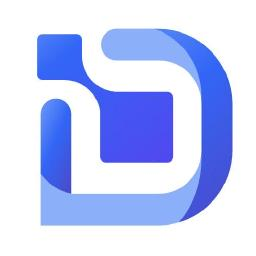}
\expandafter\def\csname modelname@mineru2-5\endcsname{MinerU2.5}
\expandafter\def\csname modelurl@mineru2-5\endcsname{https://huggingface.co/opendatalab/MinerU2.5-2509-1.2B}
\expandafter\def\csname modelicon@mineru2-5\endcsname{figures/model_icons/mineru.png}
\expandafter\def\csname modelname@monkeyocr-pro-1-2b\endcsname{MonkeyOCR-pro-1.2B}
\expandafter\def\csname modelurl@monkeyocr-pro-1-2b\endcsname{https://huggingface.co/echo840/MonkeyOCR-pro-1.2B}
\expandafter\def\csname modelicon@monkeyocr-pro-1-2b\endcsname{figures/model_icons/monkeyocr.png}
\expandafter\def\csname modelname@paddleocr-vl-1-5\endcsname{PaddleOCR-VL-1.5}
\expandafter\def\csname modelurl@paddleocr-vl-1-5\endcsname{https://huggingface.co/PaddlePaddle/PaddleOCR-VL-1.5}
\expandafter\def\csname modelicon@paddleocr-vl-1-5\endcsname{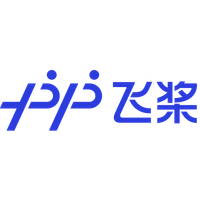}
\expandafter\def\csname modelname@glm-ocr\endcsname{GLM-OCR}
\expandafter\def\csname modelurl@glm-ocr\endcsname{https://huggingface.co/zai-org/GLM-OCR}
\expandafter\def\csname modelicon@glm-ocr\endcsname{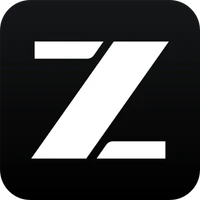}
\expandafter\def\csname modelname@openocr\endcsname{OpenOCR}
\expandafter\def\csname modelurl@openocr\endcsname{https://huggingface.co/topdu/OpenOCR}
\expandafter\def\csname modelicon@openocr\endcsname{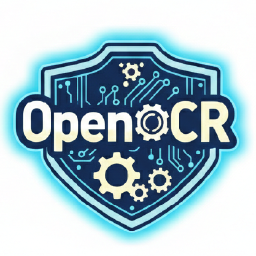}
\expandafter\def\csname modelname@olmocr-2-7b\endcsname{olmOCR-2-7B}
\expandafter\def\csname modelurl@olmocr-2-7b\endcsname{https://huggingface.co/allenai/olmOCR-2-7B-1025}
\expandafter\def\csname modelicon@olmocr-2-7b\endcsname{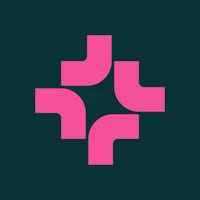}
\expandafter\def\csname modelname@olmocr-7b\endcsname{olmOCR-7B}
\expandafter\def\csname modelurl@olmocr-7b\endcsname{https://huggingface.co/allenai/olmOCR-7B-0825}
\expandafter\def\csname modelicon@olmocr-7b\endcsname{figures/model_icons/olmocr.png}
\expandafter\def\csname modelname@fd-rl\endcsname{FD-RL}
\expandafter\def\csname modelurl@fd-rl\endcsname{https://huggingface.co/DocTron/FD-RL}
\expandafter\def\csname modelicon@fd-rl\endcsname{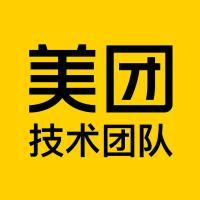}
\expandafter\def\csname modelname@logics-parsing-v2\endcsname{Logics-Parsing-v2}
\expandafter\def\csname modelurl@logics-parsing-v2\endcsname{https://huggingface.co/Logics-MLLM/Logics-Parsing-v2}
\expandafter\def\csname modelicon@logics-parsing-v2\endcsname{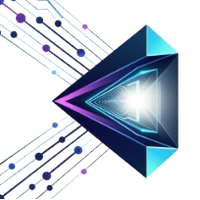}
\expandafter\def\csname modelname@ocrverse\endcsname{OCRVerse}
\expandafter\def\csname modelurl@ocrverse\endcsname{https://huggingface.co/DocTron/OCRVerse}
\expandafter\def\csname modelicon@ocrverse\endcsname{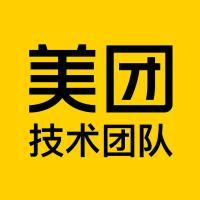}
\expandafter\def\csname modelname@qianfan-ocr\endcsname{Qianfan-OCR}
\expandafter\def\csname modelurl@qianfan-ocr\endcsname{https://huggingface.co/baidu/Qianfan-OCR}
\expandafter\def\csname modelicon@qianfan-ocr\endcsname{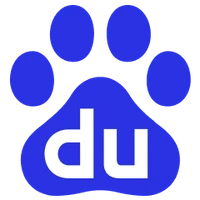}
\expandafter\def\csname modelname@nanonets-ocr2\endcsname{Nanonets-OCR2}
\expandafter\def\csname modelurl@nanonets-ocr2\endcsname{https://huggingface.co/nanonets/Nanonets-OCR2-3B}
\expandafter\def\csname modelicon@nanonets-ocr2\endcsname{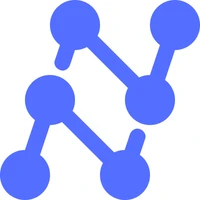}
\expandafter\def\csname modelname@deepseek-ocr-2\endcsname{DeepSeek-OCR-2}
\expandafter\def\csname modelurl@deepseek-ocr-2\endcsname{https://huggingface.co/deepseek-ai/DeepSeek-OCR-2}
\expandafter\def\csname modelicon@deepseek-ocr-2\endcsname{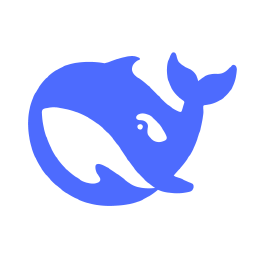}
\expandafter\def\csname modelname@ocrflux-3b\endcsname{OCRFlux-3B}
\expandafter\def\csname modelurl@ocrflux-3b\endcsname{https://huggingface.co/ChatDOC/OCRFlux-3B}
\expandafter\def\csname modelicon@ocrflux-3b\endcsname{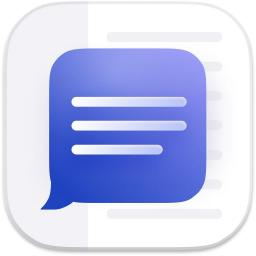}
\expandafter\def\csname modelname@deepseek-ocr\endcsname{DeepSeek-OCR}
\expandafter\def\csname modelurl@deepseek-ocr\endcsname{https://huggingface.co/deepseek-ai/DeepSeek-OCR}
\expandafter\def\csname modelicon@deepseek-ocr\endcsname{figures/model_icons/deepseek.png}
\expandafter\def\csname modelname@dots-ocr\endcsname{dots.ocr}
\expandafter\def\csname modelurl@dots-ocr\endcsname{https://huggingface.co/dots-studio/dots.ocr}
\expandafter\def\csname modelicon@dots-ocr\endcsname{figures/model_icons/dots.png}
\expandafter\def\csname modelname@firered-ocr\endcsname{FireRed-OCR}
\expandafter\def\csname modelurl@firered-ocr\endcsname{https://huggingface.co/FireRedTeam/FireRed-OCR}
\expandafter\def\csname modelicon@firered-ocr\endcsname{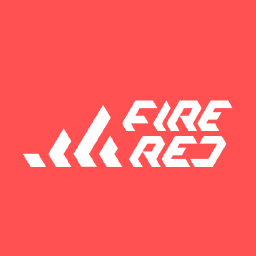}
\expandafter\def\csname modelname@hunyuanocr\endcsname{HunyuanOCR}
\expandafter\def\csname modelurl@hunyuanocr\endcsname{https://huggingface.co/tencent/HunyuanOCR/tree/main/v1.0}
\expandafter\def\csname modelicon@hunyuanocr\endcsname{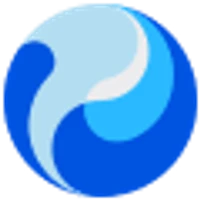}
\expandafter\def\csname modelname@unirec-0-1b\endcsname{UniRec-0.1B}
\expandafter\def\csname modelurl@unirec-0-1b\endcsname{https://huggingface.co/topdu/unirec_0_1b_onnx}
\expandafter\def\csname modelicon@unirec-0-1b\endcsname{figures/model_icons/unirec.png}
\expandafter\def\csname modelname@opendoc-0-1b\endcsname{OpenDoc-0.1B (UniRec)}
\expandafter\def\csname modelurl@opendoc-0-1b\endcsname{https://github.com/Topdu/OpenOCR/blob/main/docs/opendoc.md}
\expandafter\def\csname modelicon@opendoc-0-1b\endcsname{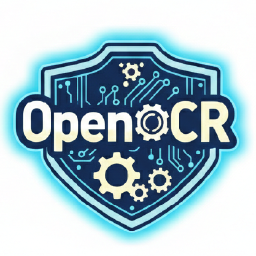}
\expandafter\def\csname modelname@qwen3-5-397b-a17b\endcsname{Qwen3.5-397B-A17B}
\expandafter\def\csname modelurl@qwen3-5-397b-a17b\endcsname{https://huggingface.co/Qwen/Qwen3.5-397B-A17B}
\expandafter\def\csname modelicon@qwen3-5-397b-a17b\endcsname{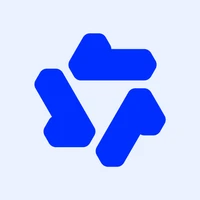}
\expandafter\def\csname modelname@qwen3-5-122b-a10b\endcsname{Qwen3.5-122B-A10B}
\expandafter\def\csname modelurl@qwen3-5-122b-a10b\endcsname{https://huggingface.co/Qwen/Qwen3.5-122B-A10B}
\expandafter\def\csname modelicon@qwen3-5-122b-a10b\endcsname{figures/model_icons/qwen.png}
\expandafter\def\csname modelname@qwen3-5-35b-a3b\endcsname{Qwen3.5-35B-A3B}
\expandafter\def\csname modelurl@qwen3-5-35b-a3b\endcsname{https://huggingface.co/Qwen/Qwen3.5-35B-A3B}
\expandafter\def\csname modelicon@qwen3-5-35b-a3b\endcsname{figures/model_icons/qwen.png}
\expandafter\def\csname modelname@qwen3-5-27b\endcsname{Qwen3.5-27B}
\expandafter\def\csname modelurl@qwen3-5-27b\endcsname{https://huggingface.co/Qwen/Qwen3.5-27B}
\expandafter\def\csname modelicon@qwen3-5-27b\endcsname{figures/model_icons/qwen.png}
\expandafter\def\csname modelname@qwen3-5-9b\endcsname{Qwen3.5-9B}
\expandafter\def\csname modelurl@qwen3-5-9b\endcsname{https://huggingface.co/Qwen/Qwen3.5-9B}
\expandafter\def\csname modelicon@qwen3-5-9b\endcsname{figures/model_icons/qwen.png}
\expandafter\def\csname modelname@qwen3-5-4b\endcsname{Qwen3.5-4B}
\expandafter\def\csname modelurl@qwen3-5-4b\endcsname{https://huggingface.co/Qwen/Qwen3.5-4B}
\expandafter\def\csname modelicon@qwen3-5-4b\endcsname{figures/model_icons/qwen.png}
\expandafter\def\csname modelname@qwen3-5-2b\endcsname{Qwen3.5-2B}
\expandafter\def\csname modelurl@qwen3-5-2b\endcsname{https://huggingface.co/Qwen/Qwen3.5-2B}
\expandafter\def\csname modelicon@qwen3-5-2b\endcsname{figures/model_icons/qwen.png}
\expandafter\def\csname modelname@qwen3-5-0-8b\endcsname{Qwen3.5-0.8B}
\expandafter\def\csname modelurl@qwen3-5-0-8b\endcsname{https://huggingface.co/Qwen/Qwen3.5-0.8B}
\expandafter\def\csname modelicon@qwen3-5-0-8b\endcsname{figures/model_icons/qwen.png}
\expandafter\def\csname modelname@qwen3-vl-8b\endcsname{Qwen3-VL-8B}
\expandafter\def\csname modelurl@qwen3-vl-8b\endcsname{https://huggingface.co/Qwen/Qwen3-VL-8B-Instruct}
\expandafter\def\csname modelicon@qwen3-vl-8b\endcsname{figures/model_icons/qwen.png}
\expandafter\def\csname modelname@qwen3-vl-4b\endcsname{Qwen3-VL-4B}
\expandafter\def\csname modelurl@qwen3-vl-4b\endcsname{https://huggingface.co/Qwen/Qwen3-VL-4B-Instruct}
\expandafter\def\csname modelicon@qwen3-vl-4b\endcsname{figures/model_icons/qwen.png}
\expandafter\def\csname modelname@qwen3-vl-2b\endcsname{Qwen3-VL-2B}
\expandafter\def\csname modelurl@qwen3-vl-2b\endcsname{https://huggingface.co/Qwen/Qwen3-VL-2B-Instruct}
\expandafter\def\csname modelicon@qwen3-vl-2b\endcsname{figures/model_icons/qwen.png}
\expandafter\def\csname modelname@kimi-k2-6\endcsname{Kimi K2.6}
\expandafter\def\csname modelurl@kimi-k2-6\endcsname{https://huggingface.co/moonshotai/Kimi-K2.6}
\expandafter\def\csname modelicon@kimi-k2-6\endcsname{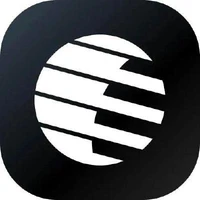}
\expandafter\def\csname modelname@step3-vl\endcsname{Step3-VL}
\expandafter\def\csname modelurl@step3-vl\endcsname{https://huggingface.co/stepfun-ai/Step3-VL-10B}
\expandafter\def\csname modelicon@step3-vl\endcsname{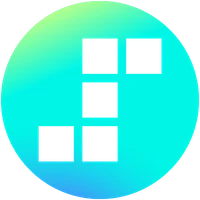}
\expandafter\def\csname modelname@minicpm-v-4-5\endcsname{MiniCPM-V-4.5}
\expandafter\def\csname modelurl@minicpm-v-4-5\endcsname{https://huggingface.co/openbmb/MiniCPM-V-4_5}
\expandafter\def\csname modelicon@minicpm-v-4-5\endcsname{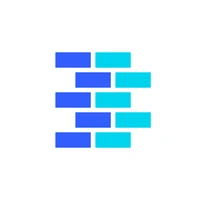}
\expandafter\def\csname modelname@gemini-3-1-pro\endcsname{Gemini-3.1-Pro}
\expandafter\def\csname modelurl@gemini-3-1-pro\endcsname{https://ai.google.dev/gemini-api/docs/models/gemini-3.1-pro-preview}
\expandafter\def\csname modelicon@gemini-3-1-pro\endcsname{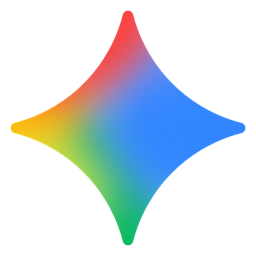}
\expandafter\def\csname modelname@claude-opus-5\endcsname{Claude Opus 5}
\expandafter\def\csname modelurl@claude-opus-5\endcsname{https://www.anthropic.com/news/claude-opus-5}
\expandafter\def\csname modelicon@claude-opus-5\endcsname{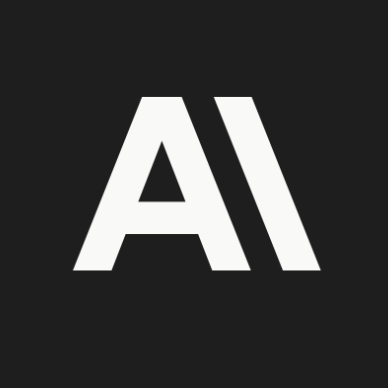}
\expandafter\def\csname modelname@deepseek-v4-flash-vision-exp\endcsname{DeepSeek-V4-Flash-Vision-Exp}
\expandafter\def\csname modelurl@deepseek-v4-flash-vision-exp\endcsname{https://huggingface.co/deepseek-ai/DeepSeek-V4-Flash-Vision-Exp}
\expandafter\def\csname modelicon@deepseek-v4-flash-vision-exp\endcsname{figures/model_icons/deepseek.png}
\expandafter\def\csname modelname@gemini-3-6-flash\endcsname{Gemini 3.6 Flash}
\expandafter\def\csname modelurl@gemini-3-6-flash\endcsname{https://ai.google.dev/gemini-api/docs/models/gemini-3.6-flash}
\expandafter\def\csname modelicon@gemini-3-6-flash\endcsname{figures/model_icons/gemini.png}
\expandafter\def\csname modelname@glm-5-3-flash\endcsname{GLM-5.3-Flash}
\expandafter\def\csname modelurl@glm-5-3-flash\endcsname{https://huggingface.co/zai-org/GLM-5.3-Flash}
\expandafter\def\csname modelicon@glm-5-3-flash\endcsname{figures/model_icons/glm.png}
\expandafter\def\csname modelname@gpt-5-6-sol\endcsname{GPT-5.6 Sol}
\expandafter\def\csname modelurl@gpt-5-6-sol\endcsname{https://developers.openai.com/api/docs/models/gpt-5.6-sol}
\expandafter\def\csname modelicon@gpt-5-6-sol\endcsname{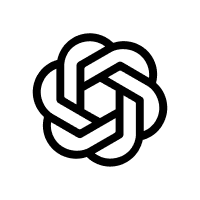}
\expandafter\def\csname modelname@minimax-m3\endcsname{MiniMax-M3}
\expandafter\def\csname modelurl@minimax-m3\endcsname{https://huggingface.co/MiniMaxAI/MiniMax-M3}
\expandafter\def\csname modelicon@minimax-m3\endcsname{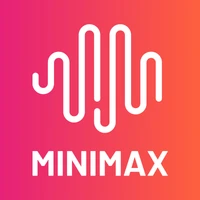}
\expandafter\def\csname modelname@qwen3-6-27b\endcsname{Qwen3.6-27B}
\expandafter\def\csname modelurl@qwen3-6-27b\endcsname{https://huggingface.co/Qwen/Qwen3.6-27B}
\expandafter\def\csname modelicon@qwen3-6-27b\endcsname{figures/model_icons/qwen.png}
\expandafter\def\csname modelname@qwen3-6-35b-a3b\endcsname{Qwen3.6-35B-A3B}
\expandafter\def\csname modelurl@qwen3-6-35b-a3b\endcsname{https://huggingface.co/Qwen/Qwen3.6-35B-A3B}
\expandafter\def\csname modelicon@qwen3-6-35b-a3b\endcsname{figures/model_icons/qwen.png}
\expandafter\def\csname modelname@qwen3-8-27b\endcsname{Qwen3.8-27B}
\expandafter\def\csname modelurl@qwen3-8-27b\endcsname{https://huggingface.co/Qwen/Qwen3.8-27B}
\expandafter\def\csname modelicon@qwen3-8-27b\endcsname{figures/model_icons/qwen.png}
\expandafter\def\csname modelname@qwen3-8-flash-next\endcsname{Qwen3.8-Flash-Next}
\expandafter\def\csname modelurl@qwen3-8-flash-next\endcsname{https://huggingface.co/Qwen/Qwen3.8-Flash-Next}
\expandafter\def\csname modelicon@qwen3-8-flash-next\endcsname{figures/model_icons/qwen.png}
\expandafter\def\csname modelname@abot-ocr\endcsname{ABot-OCR}
\expandafter\def\csname modelurl@abot-ocr\endcsname{https://huggingface.co/acvlab/ABot-OCR}
\expandafter\def\csname modelicon@abot-ocr\endcsname{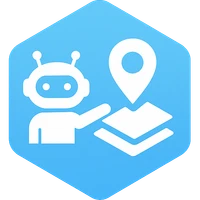}
\expandafter\def\csname modelname@hunyuanocr-1-5\endcsname{HunyuanOCR-1.5}
\expandafter\def\csname modelurl@hunyuanocr-1-5\endcsname{https://huggingface.co/tencent/HunyuanOCR}
\expandafter\def\csname modelicon@hunyuanocr-1-5\endcsname{figures/model_icons/hunyuan.png}
\expandafter\def\csname modelname@minicpm-v-4-6\endcsname{MiniCPM-V-4.6}
\expandafter\def\csname modelurl@minicpm-v-4-6\endcsname{https://huggingface.co/openbmb/MiniCPM-V-4.6}
\expandafter\def\csname modelicon@minicpm-v-4-6\endcsname{figures/model_icons/minicpm.png}
\expandafter\def\csname modelname@monkeyocr-v2-b-parsing\endcsname{MonkeyOCRv2-B-Parsing}
\expandafter\def\csname modelurl@monkeyocr-v2-b-parsing\endcsname{https://huggingface.co/zenosai/MonkeyOCRv2-B-Parsing}
\expandafter\def\csname modelicon@monkeyocr-v2-b-parsing\endcsname{figures/model_icons/monkeyocr.png}
\expandafter\def\csname modelname@ovisocr2\endcsname{OvisOCR2}
\expandafter\def\csname modelurl@ovisocr2\endcsname{https://huggingface.co/ATH-MaaS/OvisOCR2}
\expandafter\def\csname modelicon@ovisocr2\endcsname{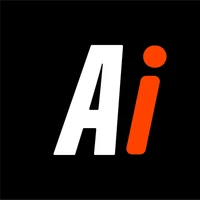}
\expandafter\def\csname modelname@paddleocr-vl-1-6-layout-corrected\endcsname{PaddleOCR-VL-1.6}
\expandafter\def\csname modelurl@paddleocr-vl-1-6-layout-corrected\endcsname{https://huggingface.co/PaddlePaddle/PaddleOCR-VL-1.6}
\expandafter\def\csname modelicon@paddleocr-vl-1-6-layout-corrected\endcsname{figures/model_icons/paddle.png}
\expandafter\def\csname modelname@teleocr\endcsname{TeleOCR}
\expandafter\def\csname modelurl@teleocr\endcsname{https://huggingface.co/StarDoc-AI/TeleOCR}
\expandafter\def\csname modelicon@teleocr\endcsname{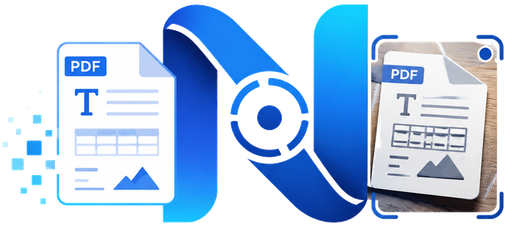}
\expandafter\def\csname modelname@unlimited-ocr\endcsname{Unlimited-OCR}
\expandafter\def\csname modelurl@unlimited-ocr\endcsname{https://huggingface.co/baidu/Unlimited-OCR}
\expandafter\def\csname modelicon@unlimited-ocr\endcsname{figures/model_icons/qianfan.png}
\expandafter\def\csname modelname@wevisdoc-4b\endcsname{WeVisDoc-4B}
\expandafter\def\csname modelurl@wevisdoc-4b\endcsname{https://huggingface.co/tencent/WeVisDoc-4B}
\expandafter\def\csname modelicon@wevisdoc-4b\endcsname{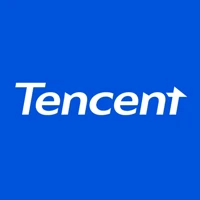}
\expandafter\def\csname modelname@wevisdoc-2b\endcsname{WeVisDoc-2B}
\expandafter\def\csname modelurl@wevisdoc-2b\endcsname{https://huggingface.co/tencent/WeVisDoc-2B}
\expandafter\def\csname modelicon@wevisdoc-2b\endcsname{figures/model_icons/tencent.png}

\newcommand{\modelentry}[1]{%
  \href{\csname modelurl@#1\endcsname}{%
    \makebox[1.35em][c]{\raisebox{-0.15ex}[0pt][0pt]{%
      \includegraphics[width=1.0em,height=1.0em,keepaspectratio]{\csname modelicon@#1\endcsname}}}%
    \hspace{0.18em}\textcolor{modelLink}{\csname modelname@#1\endcsname}}}

\newcommand{\ours}{PureDocBench}

\title{How Far Is Document Parsing from Solved?\\{\color{cyan!60!blue}\faGem}~{\color{cyan!60!blue}Pure}DocBench: A Source-Traceable Benchmark\\across Clean, Degraded, and Real-World Settings}

\author{
  {\fontsize{10}{12}\selectfont\bfseries
  Zhiheng Li$^{1,2,9,*}$ \quad
  Zongyang Ma$^{2,*}$ \quad
  Jiaxian Chen$^{3,*}$ \quad
  Jianing Zhang$^{4}$ \quad
  Zhaolong Su$^{4}$\\
  Yutong Zhang$^{5}$ \quad
  Zhiyin Yu$^{6}$ \quad
  Ruiqi Liu$^{2}$ \quad
  Xiaolei Lv$^{7}$ \quad
  Bo Li$^{7}$\\
  Jun Gao$^{7}$ \quad
  Ziqi Zhang$^{2}$ \quad
  Chunfeng Yuan$^{1,2,9,\dagger}$ \quad
  Bing Li$^{1,2,9}$ \quad
  Weiming Hu$^{1,2,8,9}$\endgraf}
  \vspace{4pt}
  {\fontsize{8.5}{10.5}\selectfont
  $^{1}$School of Artificial Intelligence, University of Chinese Academy of Sciences\\
  $^{2}$Institute of Automation, Chinese Academy of Sciences\\
  $^{3}$NWPU \quad $^{4}$JLU \quad $^{5}$USTC \quad
  $^{6}$PKU \quad $^{7}$HelloGroup \quad $^{8}$ShanghaiTech\\
  $^{9}$Beijing Key Laboratory of Super Intelligent Security of Multi-Modal Information\endgraf}
  \vspace{3pt}
  {\fontsize{8.5}{10}\selectfont $^*$Core contributors. \quad $^\dagger$Corresponding author.\\
  \texttt{$^*$lizhiheng2025@ia.ac.cn} \quad \texttt{$^\dagger$cfyuan@nlpr.ia.ac.cn}\endgraf}
}

\newtcolorbox{takeaway}{%
  colback=gray!5, colframe=black,
  boxrule=0.8pt, arc=3pt,
  left=10pt, right=10pt, top=5pt, bottom=5pt, boxsep=0pt,
  before skip=5pt, after skip=8pt,
  fontupper=\small\linespread{1.12}\selectfont}

\begin{document}
\raggedbottom
\maketitle

\begin{figure}[H]
\centering
\includegraphics[width=\textwidth]{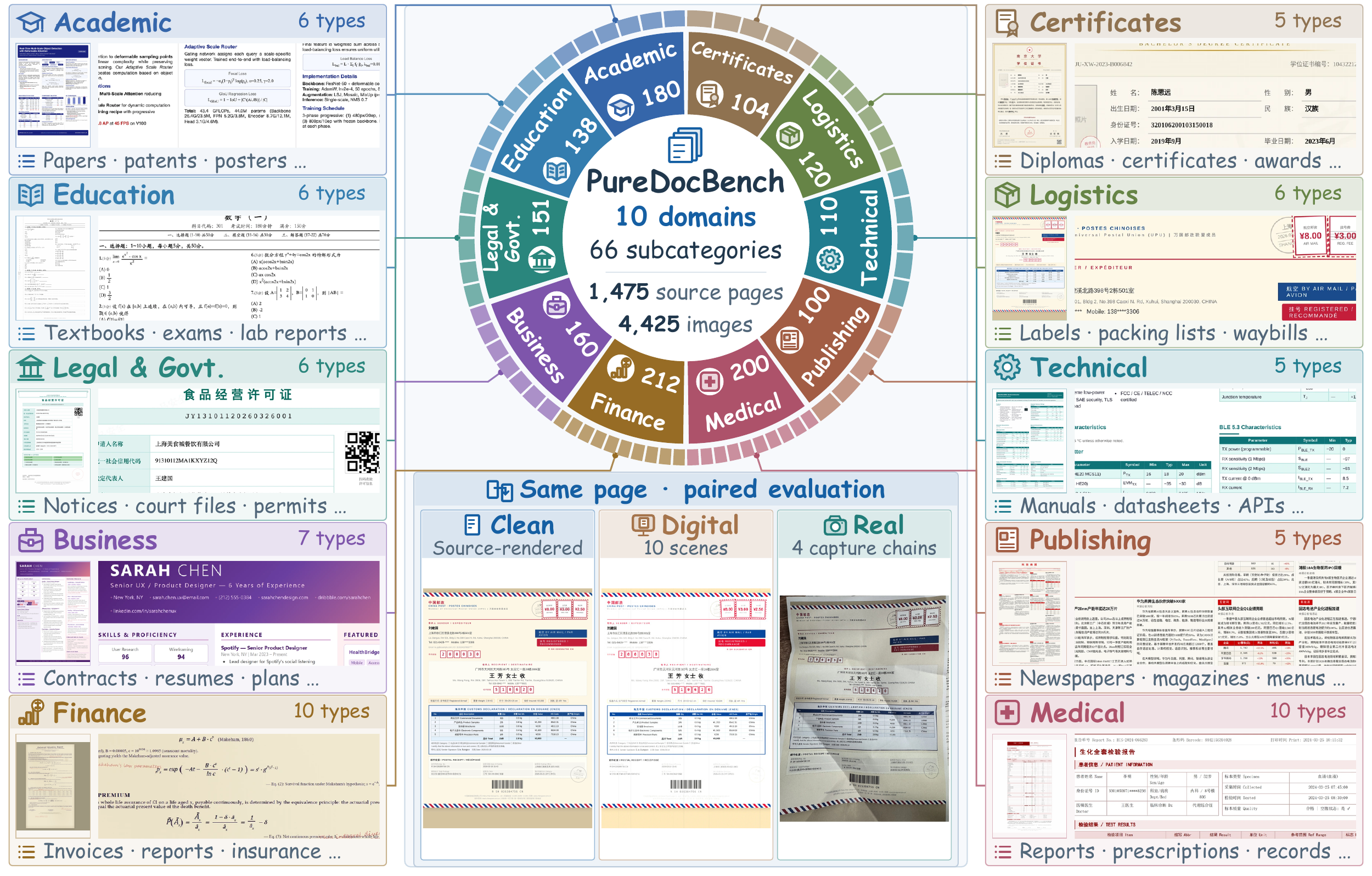}
\caption{\textbf{\ours{} overview.} The benchmark covers 10 domains and 66 subcategories, with 1{,}475 source pages and 4{,}425 matched Clean, Digital, and Real images. The three versions of each page share content and source-linked annotations for text, formulas, tables, and reading order.}
\label{fig:teaser}
\end{figure}

\begin{abstract}

Document parsing models are improving rapidly, but existing benchmarks are still insufficient for a comprehensive evaluation of their capabilities. We introduce \textbf{\ours{}}, a benchmark with \textbf{10 domains, 66 subcategories, and 1{,}475 pages}. Each page has matched Clean, Digital, and Real versions, covering clean renders, simulated degradation, and real capture or sharing conditions. The resulting 4{,}425 images have source-derived annotations for text, formulas, tables, and reading order, checked through automated validation and human review. Keeping page content and annotations fixed allows direct comparisons across image conditions. We evaluate 58 models, including pipeline-based parsers, end-to-end parsers, and general vision-language models. The best three-track average score is 79.69 out of 100, and the mean across models is 63.84. Scores drop by 3.34 points on Digital and 11.61 points on Real on average, relative to Clean. TeleOCR leads on Clean, GLM-5.3-Flash on Digital, and Gemini 3.6 Flash on Real. The benchmark dataset, source documents, generation tools, and evaluation code are publicly available at {\urlstyle{same}\url{https://github.com/zhihengli-casia/PureDocBench}}.

\end{abstract}

\section{Introduction}
\label{sec:intro}

Document parsing recovers text, table structure, formula \LaTeX{}, and reading order from document images~\citep{documentai}. Specialist models follow two broad designs. \textbf{Multi-stage systems} analyze layout before recognizing regions, using separate components (PaddleOCR-VL, GLM-OCR)~\citep{paddleocrVL,glmocr} or a shared VLM (MinerU2.5, Dolphin-v2)~\citep{mineru,minerupro,dolphinv2}. \textbf{End-to-end systems} directly map page images to document markup, as in FireRed-OCR, HunyuanOCR, WeVisDoc, and DeepSeek-OCR/-2~\citep{fireredocr,hunyuanocr,wevisdoc,deepseekocr,deepseekocr2}. General VLMs, including Qwen3-VL, Qwen3.5, and Gemini-3.1-Pro~\citep{qwen3vl,qwen35,gemini31pro}, provide another widely used approach. Rapid releases across these families increase the need for reliable evaluation across document types and acquisition conditions.

Recent benchmarks, including OmniDocBench~\citep{omnidocbench}, DocPTBench~\citep{docptbench}, and other task-focused resources~\citep{real5omnidocbench,mdpbench,ocrbenchv2,olmocr2,ocrreasoningbench}, provide valuable foundations for measuring document understanding. Collecting documents across domains is labor-intensive, and annotating their text, formulas, tables, and reading order requires substantial manual effort. Extending a benchmark to new document types or domains often means repeating the collection and annotation process.

We introduce \textbf{\ours{}} (Figure~\ref{fig:teaser}), built from generated HTML/CSS documents with source-linked annotations and matched image tracks. Its generation and annotation pipeline can be reused across document types and domains (Section~\ref{sec:construction}).

Our 58-model evaluation (Figure~\ref{fig:rankings}) yields four findings. \textbf{(1) Overall accuracy:} best Avg$_3$ 79.69, mean 63.84, and a 50.2-point best-to-worst spread. \textbf{(2) Competitive compact specialists:} MinerU2.5-Pro (1.2B) closely matches Kimi K2.6 (1T/32B active). \textbf{(3) Robustness differences:} digital/real Overall losses average $1.84$/$9.30$ for general VLMs versus $5.17$/$14.23$ for pipeline specialists. \textbf{(4) Component differences:} E2E specialists achieve the lowest mean text and reading-order errors, while general VLMs achieve the highest mean formula and table scores.

Our contributions are threefold:
\begin{itemize}[leftmargin=1.2em,itemsep=2pt,topsep=3pt,parsep=0pt]
\item \textbf{Source-traceable benchmark.} We introduce 1,475 pages across 10 domains and 66 subcategories, with source-linked annotations and matched Clean, Digital, and Real images for each page.
\item \textbf{Comprehensive evaluation.} We compare 58 specialist parsers and general VLMs across three tracks, examining overall accuracy, robustness, and performance on individual parsing components.
\item \textbf{Open resources.} We release document images, sources, annotations, generation and evaluation tools, and available model predictions for reproducible evaluation and benchmark extension.
\end{itemize}

\section{\ours{}}
\label{sec:construction}

\begin{figure}[!t]
\centering
\includegraphics[width=\textwidth]{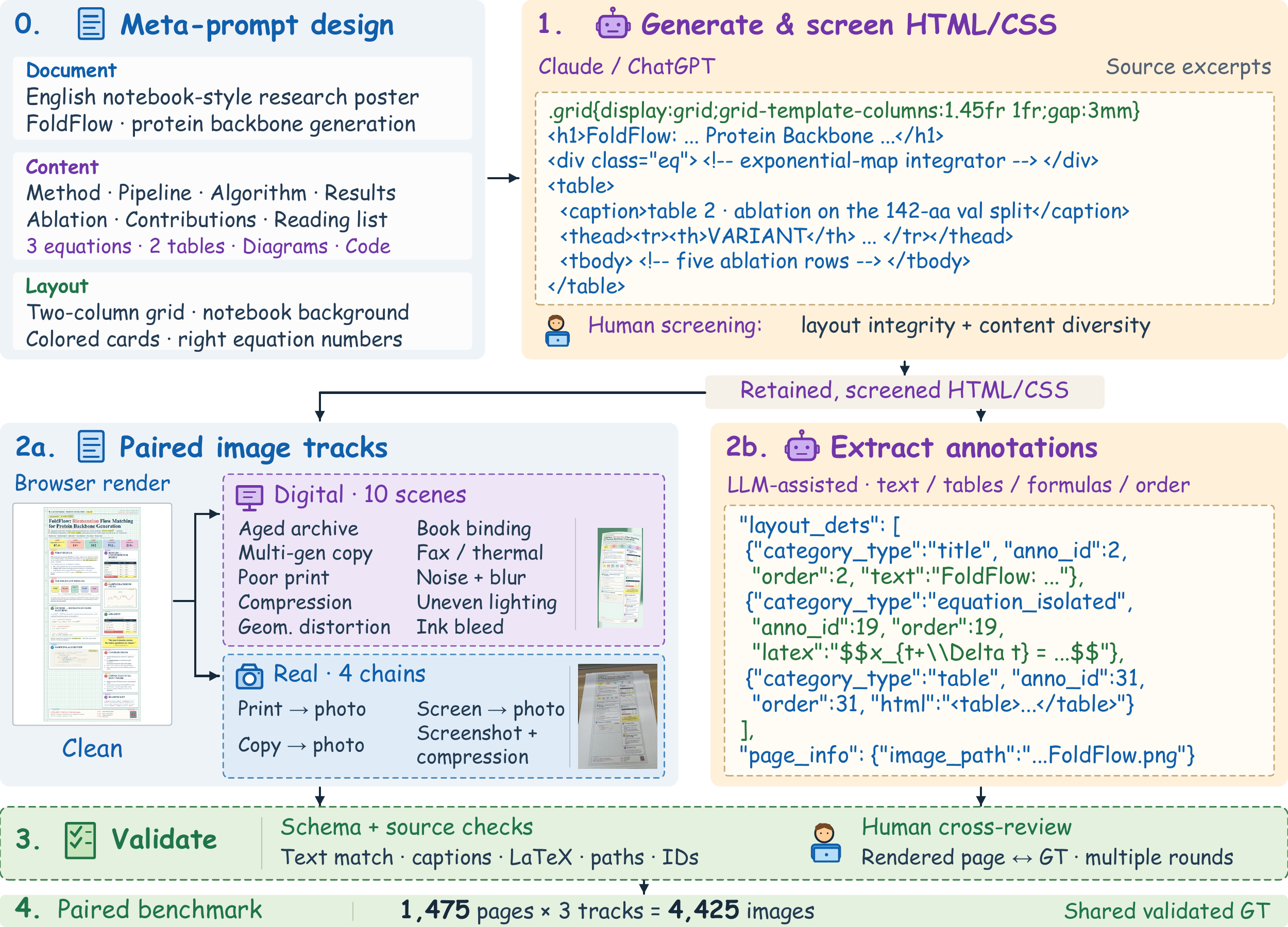}
\vspace{-8pt}
\caption{\textbf{Data construction and source-linked annotation.} Generated HTML/CSS documents provide a common source for page images and annotations of text, tables, formulas, and reading order. Matched Clean, Digital, and Real versions enable comparisons across image conditions, with automated checks and human review supporting annotation quality across all three image tracks.}
\label{fig:overview}
\end{figure}

\subsection{Data Construction}
\label{sec:taxonomy}

\ours{} links rendered images and annotations to shared HTML/CSS sources (Figure~\ref{fig:overview}). Compared with collecting and annotating public pages, generating document content from scratch and rendering images directly helps reduce the risk of overlap with model training data. Newly generated content and source-traceable annotations motivate \emph{Pure}.

\paragraph{Taxonomy (Figure~\ref{fig:teaser}).} Ten domains and 66 subcategories cover financial invoices, medical records, legal contracts, and logistics documents~\citep{idp_fortune2026,idp_gvr2025}.

\paragraph{Paired image tracks (2a).} Each of 1,475 pages has clean, digitally degraded, and real-degraded versions: 4,425 images sharing GT. Ten digital scene templates include aging, binding, photocopying, compression, lighting, and geometric artifacts. Four real-track chains comprise phone capture of prints, phone capture of photocopies, screen photography, and screenshots with social-media compression. Pairing fixes content while varying acquisition; Appendix~\ref{app:degradation_params} lists operations and weights.

\paragraph{Source and annotation pipeline (0--4).} Subcategory prompts specify document type, layout, domain, language, and element composition. LLMs emit self-contained HTML/CSS with text, structured tables, and \LaTeX{} formulas; human screening rejects collapsed layouts and repetitive content. A browser renders high-resolution PNGs. LLM-assisted extraction reads the same HTML to produce OmniDocBench-format text, table HTML, formulas, and reading order. Deterministic schema/source checks and multiple human cross-review rounds validate all 1,475 GT pages.

Retained sources support annotation inspection and regeneration and remain hidden from evaluated models. New prompts extend pages and document types through the same validation/review pipeline, enabling refreshed future splits.

\begingroup
\definecolor{benchInk}{HTML}{243E55}
\definecolor{benchLink}{HTML}{245BA3}
\definecolor{benchMuted}{HTML}{69798B}
\definecolor{benchRule}{HTML}{96B5CB}
\definecolor{benchStripe}{HTML}{EEF3FA}
\definecolor{benchHeader}{HTML}{244766}
\definecolor{benchHeaderLight}{HTML}{DFEAF6}
\definecolor{benchBlueBand}{HTML}{CDDEF5}
\definecolor{benchBlueInk}{HTML}{24559A}
\definecolor{benchVioletBand}{HTML}{E2D8F2}
\definecolor{benchVioletInk}{HTML}{64438E}
\definecolor{benchGreenBand}{HTML}{CDE7E1}
\definecolor{benchGreenInk}{HTML}{176A62}
\definecolor{benchTeal}{HTML}{B7E2D5}
\definecolor{benchCheck}{HTML}{008777}
\definecolor{benchPartialColor}{HTML}{B9781E}
\definecolor{benchNoColor}{HTML}{8B98A5}
\expandafter\def\csname benchurl@im2latex\endcsname{https://github.com/harvardnlp/im2markup}
\expandafter\def\csname benchurl@publaynet\endcsname{https://github.com/ibm-aur-nlp/PubLayNet}
\expandafter\def\csname benchurl@docbank\endcsname{https://github.com/doc-analysis/DocBank}
\expandafter\def\csname benchurl@tablebank\endcsname{https://github.com/doc-analysis/TableBank}
\expandafter\def\csname benchurl@teds\endcsname{https://github.com/ibm-aur-nlp/PubTabNet}
\expandafter\def\csname benchurl@doclaynet\endcsname{https://github.com/DS4SD/DocLayNet}
\expandafter\def\csname benchurl@pubtables1m\endcsname{https://github.com/microsoft/table-transformer}
\expandafter\def\csname benchurl@docvqa\endcsname{https://www.docvqa.org/datasets/docvqa}
\expandafter\def\csname benchurl@ocrbench\endcsname{https://github.com/Yuliang-Liu/MultimodalOCR/tree/main/OCRBench}
\expandafter\def\csname benchurl@ocrbenchv2\endcsname{https://github.com/Yuliang-Liu/MultimodalOCR/tree/main/OCRBench_v2}
\expandafter\def\csname benchurl@ccocr\endcsname{https://github.com/AlibabaResearch/AdvancedLiterateMachinery/tree/main/Benchmarks/CC-OCR}
\expandafter\def\csname benchurl@dogr\endcsname{https://github.com/Tencent/DOGR}
\expandafter\def\csname benchurl@cosyn\endcsname{https://yueyang.ai/cosyn/}
\expandafter\def\csname benchurl@madqa\endcsname{https://github.com/OxRML/MADQA}
\expandafter\def\csname benchurl@synthdocbench\endcsname{https://github.com/ServiceNow/SynthDocBench}
\expandafter\def\csname benchurl@readoc\endcsname{https://github.com/icip-cas/READoc}
\expandafter\def\csname benchurl@omnidocbench\endcsname{https://github.com/opendatalab/OmniDocBench/tree/v1_5}
\expandafter\def\csname benchurl@olmocr\endcsname{https://github.com/allenai/olmocr/blob/main/olmocr/bench/README.md}
\expandafter\def\csname benchurl@docptbench\endcsname{https://github.com/Topdu/DocPTBench}
\expandafter\def\csname benchurl@real5omnidocbench\endcsname{https://huggingface.co/datasets/PaddlePaddle/Real5-OmniDocBench}
\expandafter\def\csname benchurl@mdpbench\endcsname{https://github.com/Yuliang-Liu/MultimodalOCR/tree/main/MDPBench}
\expandafter\def\csname benchurl@omniparsingbench\endcsname{https://arxiv.org/abs/2603.09677}
\expandafter\def\csname benchurl@parsebench\endcsname{https://github.com/run-llama/ParseBench}
\expandafter\def\csname benchurl@morebench\endcsname{https://github.com/zimoqingfeng/MORE}
\expandafter\def\csname benchurl@mpdocbenchparse\endcsname{https://github.com/Tongyi-Zhiwen/Qwen-Doc/tree/main/MPDocBench}
\expandafter\def\csname benchurl@drdocbench\endcsname{https://github.com/2077AI/DrDocBench}
\expandafter\def\csname benchurl@puredocbench\endcsname{https://github.com/zhihengli-casia/PureDocBench}
\expandafter\def\csname benchurl@ccocrv2\endcsname{https://github.com/eioss/CC-OCR-V2}

\hypersetup{citecolor=benchMuted,urlcolor=benchLink}
\newcommand{\benchref}[2]{%
  \href{\csname benchurl@#2\endcsname}{\textcolor{benchLink}{#1}}%
  {\textcolor{benchMuted}{\,\citeyearpar{#2}}}}
\newcommand{\benchYes}{\textcolor{benchCheck}{\ding{51}}}
\newcommand{\benchPartial}{\textcolor{benchPartialColor}{\faAdjust}}
\newcommand{\benchNo}{\textcolor{benchNoColor}{---}}
\newcommand{\benchSrc}{\textcolor{benchVioletInk}{\faCode}}
\newcommand{\benchModel}{\textcolor{benchVioletInk}{\faRobot}}
\newcommand{\benchHuman}{\textcolor{benchVioletInk}{\faUserCheck}}
\newcommand{\benchInherited}{\textcolor{benchVioletInk}{\faCopy}}
\newcommand{\benchgroup}[4]{%
  \multicolumn{9}{@{}l@{}}{\cellcolor{#1}\rule{0pt}{11pt}%
  \parbox{\textwidth}{\hspace{3pt}\textcolor{#2}{#3\quad\textbf{#4}}}}\\[2pt]}
\newcolumntype{Z}[1]{>{\centering\arraybackslash}m{#1}}
\newcommand{\benchhead}[2]{\makecell{\textcolor{benchLink}{#1}\\#2}}
\begin{table}[t]
\centering
\caption{\textbf{Comparison of document benchmarks and corpora.} Resources span component recognition, document understanding, and full-page parsing. \ours{} combines four parsing targets, source-linked annotations, and matched digital and physical image variants across 10 domains.}
\label{tab:benchmark_comparison}
\vspace{2pt}
\rmfamily\fontsize{8}{9.4}\selectfont
\setlength{\tabcolsep}{1.5pt}
\renewcommand{\arraystretch}{1.08}
\arrayrulecolor{benchRule}
\begin{tabular}{@{}L{113pt}Z{47pt}Z{24pt}Z{28pt}Z{32pt}Z{32pt}Z{36pt}Z{30pt}Z{31pt}@{}}
\toprule[.7pt]
\rowcolor{benchHeader}
\rule[-5pt]{0pt}{20pt}\textcolor{white}{\textbf{Resource}} & \textcolor{white}{\textbf{Corpus size}} & \multicolumn{4}{Z{125pt}}{\textcolor{white}{\textbf{Parsing targets}}} & \shortstack{\textcolor{white}{\textbf{Label}}\\\textcolor{white}{\textbf{source}}} & \multicolumn{2}{Z{64pt}@{}}{\shortstack{\textcolor{white}{\textbf{Paired image}}\\\textcolor{white}{\textbf{variants}}}} \\
\cmidrule(lr{2pt}){3-6}\cmidrule(l{2pt}){8-9}
\rowcolor{benchHeaderLight}
& & \benchhead{\faAlignLeft}{Text} & \benchhead{\faTable}{Tables} & \benchhead{\faSuperscript}{Formulas} & \benchhead{\faSortAmountDown}{Reading\\order} & & \benchhead{\faDesktop}{Digital} & \benchhead{\faCamera}{Physical} \\
\midrule[.5pt]
\benchgroup{benchBlueBand}{benchBlueInk}{\faLayerGroup}{Component recognition}
\benchref{im2latex-100k}{im2latex} & 104k formulas & \benchNo & \benchNo & \benchYes & \benchNo & \benchSrc & \benchNo & \benchNo \\[4pt]
\rowcolor{benchStripe}
\benchref{PubLayNet}{publaynet} & 360k+ pages & \benchNo & \benchNo & \benchNo & \benchNo & \benchSrc & \benchNo & \benchNo \\[4pt]
\benchref{DocBank}{docbank} & 500k pages & \benchNo & \benchNo & \benchNo & \benchNo & \benchSrc & \benchNo & \benchNo \\[4pt]
\rowcolor{benchStripe}
\benchref{TableBank}{tablebank} & 417k tables & \benchNo & \benchYes & \benchNo & \benchNo & \benchSrc & \benchNo & \benchNo \\[4pt]
\benchref{PubTabNet}{teds} & 568k+ tables & \benchPartial & \benchYes & \benchNo & \benchNo & \benchSrc & \benchNo & \benchNo \\[4pt]
\rowcolor{benchStripe}
\benchref{DocLayNet}{doclaynet} & 80,863 pages & \benchNo & \benchNo & \benchNo & \benchNo & \benchHuman & \benchNo & \benchNo \\[4pt]
\benchref{PubTables-1M}{pubtables1m} & 948k tables & \benchNo & \benchYes & \benchNo & \benchNo & \benchSrc & \benchNo & \benchNo \\[4pt]
\benchgroup{benchVioletBand}{benchVioletInk}{\faComments}{Document understanding and synthetic supervision}
\benchref{DocVQA}{docvqa} & 12,767 images & \benchNo & \benchNo & \benchNo & \benchNo & \benchHuman & \benchNo & \benchNo \\[4pt]
\rowcolor{benchStripe}
\benchref{OCRBench}{ocrbench} & 1k QA pairs & \benchYes & \benchNo & \benchPartial & \benchNo & \benchInherited\,\benchHuman & \benchNo & \benchNo \\[4pt]
\benchref{OCRBench v2}{ocrbenchv2} & 10k QA pairs & \benchYes & \benchYes & \benchYes & \benchNo & \benchInherited\,\benchHuman & \benchNo & \benchNo \\[4pt]
\rowcolor{benchStripe}
\benchref{CC-OCR}{ccocr} & 7,058 images & \benchYes & \benchYes & \benchYes & \benchNo & \benchInherited\,\benchModel\,\benchHuman & \benchNo & \benchNo \\[4pt]
\benchref{CC-OCR v2}{ccocrv2} & 7,093 samples & \benchYes & \benchYes & \benchYes & \benchNo & \benchInherited\,\benchModel\,\benchHuman & \benchNo & \benchNo \\[4pt]
\rowcolor{benchStripe}
\benchref{DOGR-Bench}{dogr} & 3.6k samples & \benchNo & \benchNo & \benchNo & \benchNo & \benchSrc\,\benchModel & \benchNo & \benchNo \\[4pt]
\benchref{CoSyn-400K}{cosyn} & 400k images & \benchNo & \benchNo & \benchNo & \benchNo & \benchSrc\,\benchModel & \benchNo & \benchNo \\[4pt]
\rowcolor{benchStripe}
\benchref{MADQA}{madqa} & 800 PDFs & \benchNo & \benchNo & \benchNo & \benchNo & \benchHuman & \benchNo & \benchNo \\[4pt]
\benchref{SynthDocBench}{synthdocbench} & 200 docs & \benchNo & \benchNo & \benchNo & \benchNo & \benchSrc\,\benchModel\,\benchHuman & \benchNo & \benchNo \\[4pt]
\benchgroup{benchGreenBand}{benchGreenInk}{\faIcon{file-alt}}{Full-document parsing}
\benchref{READoc}{readoc} & 3,576 docs & \benchYes & \benchYes & \benchYes & \benchYes & \benchSrc & \benchNo & \benchNo \\[4pt]
\rowcolor{benchStripe}
\benchref{OmniDocBench v1.5}{omnidocbench} & 1,355 pages & \benchYes & \benchYes & \benchYes & \benchYes & \benchModel\,\benchHuman & \benchNo & \benchNo \\[4pt]
\benchref{olmOCR-bench}{olmocr} & 1,403 pages & \benchYes & \benchPartial & \benchYes & \benchYes & \benchSrc\,\benchModel\,\benchHuman & \benchNo & \benchNo \\[4pt]
\rowcolor{benchStripe}
\benchref{DocPTBench}{docptbench} & 981 base pages & \benchYes & \benchYes & \benchYes & \benchYes & \benchInherited & \benchYes & \benchPartial \\[4pt]
\benchref{Real5-OmniDocBench}{real5omnidocbench} & 1,355 $\times$ 5 & \benchYes & \benchYes & \benchYes & \benchYes & \benchInherited & \benchNo & \benchYes \\[4pt]
\rowcolor{benchStripe}
\benchref{MDPBench}{mdpbench} & 850 $\times$ 4 & \benchYes & \benchYes & \benchYes & \benchYes & \benchModel\,\benchHuman & \benchNo & \benchYes \\[4pt]
\benchref{LogicsDocBench}{omniparsingbench} & 900 pages & \benchYes & \benchYes & \benchYes & \benchYes & \benchNo & \benchNo & \benchNo \\[4pt]
\rowcolor{benchStripe}
\benchref{ParseBench}{parsebench} & 2,078 pages & \benchYes & \benchYes & \benchPartial & \benchYes & \benchModel\,\benchHuman & \benchNo & \benchNo \\[4pt]
\benchref{MORE}{morebench} & $\sim$1.3k images & \benchYes & \benchYes & \benchYes & \benchYes & \benchModel\,\benchHuman & \benchNo & \benchNo \\[4pt]
\rowcolor{benchStripe}
\benchref{MPDocBench-Parse}{mpdocbenchparse} & 3,246 pages & \benchYes & \benchYes & \benchYes & \benchYes & \benchModel\,\benchHuman & \benchNo & \benchNo \\[4pt]
\benchref{Dr. DocBench}{drdocbench} & 4,514 pages & \benchYes & \benchYes & \benchYes & \benchYes & \benchModel\,\benchHuman & \benchNo & \benchNo \\[4pt]
\midrule[.5pt]
\rowcolor{benchTeal}
\href{\csname benchurl@puredocbench\endcsname}{\textcolor{benchLink}{\faGem\,\textbf{\ours{}}}}\,\textbf{(Ours)} & \textbf{1,475 $\times$ 3} & \benchYes & \benchYes & \benchYes & \benchYes & \benchSrc\,\benchModel\,\benchHuman & \benchYes & \benchYes \\[4pt]
\bottomrule[.7pt]
\end{tabular}
\vspace{3pt}
\parbox{\textwidth}{\fontsize{7}{8.3}\selectfont
\benchYes\ reported; \benchPartial\ restricted scope; \benchNo\ unreported.
\quad\benchSrc\ source markup; \benchModel\ model; \benchHuman\ human labels/review; \benchInherited\ inherited labels.\\
\textcolor{benchMuted}{Size: original corpus units; $\times$ = base pages $\times$ views (Real5: captures only; MDPBench/PureDocBench: original included).}}
\vspace{-5pt}
\end{table}
\arrayrulecolor{black}
\endgroup

\subsection{Comparison with Related Benchmarks}
\label{sec:comparison}

Table~\ref{tab:benchmark_comparison} organizes benchmarks and related corpora by task; Appendix~\ref{app:benchmark_comparison} records comparison definitions and version-specific evidence.

\paragraph{Component recognition and source supervision.} The component datasets in Table~\ref{tab:benchmark_comparison} cover layout, table structure, and printed formulas, establishing PDF--XML, Word, and \LaTeX{} supervision. FinTabNet~\citep{fintabnet} adds financial tables; CROHME~\citep{crohme2011}, ReadingBank~\citep{layoutreader}, and HRDoc~\citep{hrdoc} address handwritten mathematics, reading order, and document hierarchy. FUNSD, SROIE, and CORD~\citep{funsd,sroie,cord} target form/receipt fields. Source-derived supervision also underlies Nougat~\citep{nougat} and READoc; CoSyn~\citep{cosyn} adds code-generated text-rich instruction-tuning data.

\paragraph{Document understanding.} DocVQA~\citep{docvqa}, InfographicVQA~\citep{infographicvqa}, DUDE~\citep{dude}, and MMLongBench-Doc~\citep{mmlongbenchdoc} evaluate answers grounded in documents. SynthDocBench~\citep{synthdocbench} uses controlled synthetic long documents. OCRBench/v2 and CC-OCR/v2~\citep{ocrbench,ocrbenchv2,ccocr,ccocrv2} probe broader OCR capabilities, including parsing subtasks; OCR-Reasoning~\citep{ocrreasoningbench} tests text-rich reasoning, MMDocBench~\citep{mmdocbench} adds fine-grained perception, DOGR-Bench~\citep{dogr} targets grounding and referring, and MADQA~\citep{madqa} studies agentic search over document collections. These task-specific evaluations complement full-document structured recovery.

\paragraph{Full-document parsing.} Table~\ref{tab:benchmark_comparison} spans structured extraction, targeted unit tests, and semantic utility. DocPTBench, Real5-OmniDocBench, and MDPBench provide matched acquisition stress, while MDPBench and MORE broaden language coverage. MPDocBench-Parse and Dr.~DocBench extend evaluation to document hierarchy and expert content; FinixDocBench~\citep{finixdoc} adds financial workflows and camera-captured inputs. Building on this lineage, \ours{} combines freshly generated, retained HTML/CSS sources with human-reviewed annotations and matched clean/digital/real images across 10 domains. Authentic-document and multi-page benchmarks provide complementary coverage.

\FloatBarrier
\section{Experiments}
\label{sec:experiments}

\subsection{Setup}
\label{sec:setup}

\paragraph{Metrics.} We adopt the OmniDocBench evaluation framework~\citep{omnidocbench}: TextEdit$\downarrow$ (page-level normalized edit distance), FormulaCDM$\uparrow$~\citep{cdm} (render-based character detection matching for formulas), TableTEDS$\uparrow$~\citep{teds} (tree-edit-distance similarity on table HTML), and ReadOrderEdit$\downarrow$ (edit distance on reading order). Per-track \textbf{Overall}$\uparrow$ is the mean of TextEdit, FormulaCDM, and TableTEDS (ROEdit is reported separately): $\mathrm{Overall} = \big((1{-}\mathrm{TextEdit}) \times 100 + \mathrm{FormulaCDM} + \mathrm{TableTEDS}\big)/3$. \textbf{Avg}$_3$ is the mean of the three track Overalls (Clean / Digital / Real).

\paragraph{Models (58).} We prioritize recent releases and compare \textbf{13 pipeline / multi-stage specialists}, \textbf{19 end-to-end specialists}, and \textbf{26 general-purpose VLMs} (Table~\ref{tab:leaderboard}). Pipeline specialists include PaddleOCR-VL-1.6, MinerU2.5-Pro, MonkeyOCRv2-B-Parsing, and TeleOCR; end-to-end specialists include OvisOCR2, WeVisDoc-2B/4B, and HunyuanOCR-1.5. General VLMs span the Qwen and MiniCPM families alongside Claude Opus~5, Gemini~3.6~Flash, GLM-5.3-Flash, GPT-5.6~Sol, and Kimi~K2.6. These models cover releases through September 2026 and multiple parameter scales.

\begin{figure}[!t]
\centering
\input{data/ranking_links.tex}
\caption{\textbf{Accuracy and robustness of 58 models.} Three-track mean accuracy (Avg$_3$) and losses relative to Clean reveal differences in parsing accuracy and robustness across the three architectural families. Losses are Clean$-$Digital and Clean$-$Real in points; negative values indicate gains.}
\label{fig:rankings}
\end{figure}

\subsection{Main Results}
\label{sec:main_results}

\nocite{deepseekocr,deepseekocr2,dolphinv2,dotsmocr,dotsocr,fdrl,fireredocr,gemini31pro,glmocr,hunyuanocr,kimik26,logicsparsingv2,mineru,minerupro,minicpmv,monkeyocr,nanonetsocr2,ocrflux,ocrverse,olmocr,olmocr2,opendoc,openocr,paddleocrVL,qianfanocr,step3vl,unirec,youtuparsing}

\textbf{Document parsing still has room for improvement.} Across 58 models, the best Avg$_3$ is 79.69 out of 100 (GLM-5.3-Flash), the model mean is 63.84, and the best-to-worst spread is 50.2 points. The median Avg$_3$ is 66.20. Only seven models exceed 75, while 41 of the 58 models score below 70.

\textbf{Compact specialist parsers remain competitive.} MinerU2.5-Pro (1.2B, Avg$_3$ 70.07) closely matches Kimi K2.6 (1T total / 32B active, 70.10) and Gemini-3.1-Pro (70.43). WeVisDoc-4B achieves an Avg$_3$ of 75.57, exceeding Qwen3.5-122B-A10B (74.11) by 1.46 points. WeVisDoc-2B scores 73.69, outperforming Qwen3.5-27B (69.57) by 4.12 points with only 7.4\% of the parameters. FD-RL and Logics-Parsing-v2, both 4B specialists, score 73.92 and 72.61, surpassing Kimi K2.6 by 3.82 and 2.51 points, respectively.

\textbf{Degradation changes model rankings.} Across all 58 models, mean Overall losses are $3.34$/$11.61$ points under digital/real degradation. TeleOCR leads Clean (87.16), GLM-5.3-Flash leads Digital (81.03) and Avg$_3$ (79.69), and Gemini 3.6 Flash rises from Clean rank 6 to Real rank 1 (75.73). By architecture, Pipeline specialists lose $5.17$/$14.23$ points, E2E specialists $4.14$/$12.99$, and general VLMs $1.84$/$9.30$ on average.

\begin{table}[!tp]
\centering
\caption{\textbf{Three-track parsing accuracy for 58 models across architectures.} Clean, Digital, and Real report per-track Overall on the same 1,475 source pages; the final column is their mean (Avg$_3$). Colored changes are relative to Clean (\textcolor{red!75!black}{decrease}/\textcolor{green!45!black}{increase}). \textbf{Bold}/\underline{underline}: best/runner-up.}
\label{tab:leaderboard}
{\fontsize{9}{10}\selectfont
\definecolor{resultDrop}{HTML}{B52F3F}
\definecolor{resultGain}{HTML}{167653}
\newcommand{\resulttrack}[2]{\makebox[23pt][r]{#1}\hspace{3pt}{\fontsize{8}{9}\selectfont\makebox[35pt][r]{#2}}}
\newlength{\resultnumberwidth}
\settowidth{\resultnumberwidth}{00.00}
\newcommand{\resultheader}[2]{\makebox[\resultnumberwidth][c]{#1}\hspace*{#2}}
\setlength{\tabcolsep}{2pt}
\renewcommand{\arraystretch}{0.95}
\begin{tabular}{>{\raggedright\arraybackslash}p{\dimexpr\textwidth-193pt-10\tabcolsep\relax}>{\raggedleft\arraybackslash}p{28pt}*{2}{>{\raggedleft\arraybackslash}p{64pt}}>{\raggedleft\arraybackslash}p{37pt}}
\toprule
\textbf{Model (release)} & \resultheader{\textcolor{cleanInk}{\textbf{Clean}}$\uparrow$}{7pt} & \resultheader{\textcolor{digitalInk}{\textbf{Digital}}$\uparrow$}{38pt} & \resultheader{\textcolor{realInk}{\textbf{Real}}$\uparrow$}{38pt} & \resultheader{\textbf{Overall}$\uparrow$}{7pt} \\
\midrule
\rowcolor{pipelineBand}\multicolumn{5}{l}{\strut\textbf{Pipeline / multi-stage specialists}} \\
\rowcolor{resultStripe}
\modelentry{paddleocr-vl-1-6-layout-corrected}\nobreak\hspace{0.3em}{\fontsize{8.5}{9}\selectfont \href{https://huggingface.co/PaddlePaddle/PaddleOCR-VL-1.6}{\textcolor{black!65}{(2026-05)}}} & 64.46\hspace*{7pt} & \resulttrack{59.29}{\textcolor{resultDrop}{\ensuremath{(-5.17)}}} & \resulttrack{54.24}{\textcolor{resultDrop}{\ensuremath{(-10.22)}}} & 59.33\hspace*{7pt} \\
\modelentry{youtu-parsing}\nobreak\hspace{0.3em}{\fontsize{8.5}{9}\selectfont \href{https://arxiv.org/html/2601.20430v1}{\textcolor{black!65}{(2026-01)}}} & 75.02\hspace*{7pt} & \resulttrack{69.66}{\textcolor{resultDrop}{\ensuremath{(-5.36)}}} & \resulttrack{60.29}{\textcolor{resultDrop}{\ensuremath{(-14.73)}}} & 68.32\hspace*{7pt} \\
\rowcolor{resultStripe}
\modelentry{dotsmocr}\nobreak\hspace{0.3em}{\fontsize{8.5}{9}\selectfont \href{https://github.com/Yuliang-Liu/MonkeyOCR}{\textcolor{black!65}{(2026-03)}}} & 76.27\hspace*{7pt} & \resulttrack{73.16}{\textcolor{resultDrop}{\ensuremath{(-3.11)}}} & \resulttrack{61.73}{\textcolor{resultDrop}{\ensuremath{(-14.54)}}} & 70.39\hspace*{7pt} \\
\modelentry{glm-ocr}\nobreak\hspace{0.3em}{\fontsize{8.5}{9}\selectfont \href{https://docs.z.ai/release-notes/new-released}{\textcolor{black!65}{(2026-02)}}} & 68.65\hspace*{7pt} & \resulttrack{63.06}{\textcolor{resultDrop}{\ensuremath{(-5.59)}}} & \resulttrack{58.31}{\textcolor{resultDrop}{\ensuremath{(-10.34)}}} & 63.34\hspace*{7pt} \\
\rowcolor{resultStripe}
\modelentry{mineru2-5-pro}\nobreak\hspace{0.3em}{\fontsize{8.5}{9}\selectfont \href{https://opendatalab.github.io/MinerU/reference/changelog/}{\textcolor{black!65}{(2026-04)}}} & 75.87\hspace*{7pt} & \resulttrack{71.77}{\textcolor{resultDrop}{\ensuremath{(-4.10)}}} & \resulttrack{62.56}{\textcolor{resultDrop}{\ensuremath{(-13.31)}}} & 70.07\hspace*{7pt} \\
\modelentry{mineru2-5}\nobreak\hspace{0.3em}{\fontsize{8.5}{9}\selectfont \href{https://opendatalab.github.io/MinerU/reference/changelog/}{\textcolor{black!65}{(2025-09)}}} & 74.90\hspace*{7pt} & \resulttrack{68.92}{\textcolor{resultDrop}{\ensuremath{(-5.98)}}} & \resulttrack{59.15}{\textcolor{resultDrop}{\ensuremath{(-15.75)}}} & 67.66\hspace*{7pt} \\
\rowcolor{resultStripe}
\modelentry{monkeyocr-v2-b-parsing}\nobreak\hspace{0.3em}{\fontsize{8.5}{9}\selectfont \href{https://huggingface.co/zenosai/MonkeyOCRv2-B-Parsing}{\textcolor{black!65}{(2026-07)}}} & 69.82\hspace*{7pt} & \resulttrack{66.08}{\textcolor{resultDrop}{\ensuremath{(-3.74)}}} & \resulttrack{58.83}{\textcolor{resultDrop}{\ensuremath{(-10.99)}}} & 64.91\hspace*{7pt} \\
\modelentry{monkeyocr-pro-3b}\nobreak\hspace{0.3em}{\fontsize{8.5}{9}\selectfont \href{https://github.com/Yuliang-Liu/MonkeyOCR/issues/252#issuecomment-3216586234}{\textcolor{black!65}{(2025-08)}}} & 62.23\hspace*{7pt} & \resulttrack{57.40}{\textcolor{resultDrop}{\ensuremath{(-4.83)}}} & \resulttrack{46.49}{\textcolor{resultDrop}{\ensuremath{(-15.74)}}} & 55.37\hspace*{7pt} \\
\rowcolor{resultStripe}
\modelentry{monkeyocr-pro-1-2b}\nobreak\hspace{0.3em}{\fontsize{8.5}{9}\selectfont \href{https://huggingface.co/echo840/MonkeyOCR-pro-1.2B}{\textcolor{black!65}{(2025-07)}}} & 61.09\hspace*{7pt} & \resulttrack{55.72}{\textcolor{resultDrop}{\ensuremath{(-5.37)}}} & \resulttrack{43.82}{\textcolor{resultDrop}{\ensuremath{(-17.27)}}} & 53.54\hspace*{7pt} \\
\modelentry{opendoc-0-1b}\nobreak\hspace{0.3em}{\fontsize{8.5}{9}\selectfont \href{https://github.com/Topdu/OpenOCR}{\textcolor{black!65}{(2025-12)}}} & 60.28\hspace*{7pt} & \resulttrack{52.46}{\textcolor{resultDrop}{\ensuremath{(-7.82)}}} & \resulttrack{44.27}{\textcolor{resultDrop}{\ensuremath{(-16.01)}}} & 52.34\hspace*{7pt} \\
\rowcolor{resultStripe}
\modelentry{openocr}\nobreak\hspace{0.3em}{\fontsize{8.5}{9}\selectfont \href{https://huggingface.co/topdu/OpenOCR}{\textcolor{black!65}{(2024-11)}}} & 32.70\hspace*{7pt} & \resulttrack{30.03}{\textcolor{resultDrop}{\ensuremath{(-2.67)}}} & \resulttrack{25.73}{\textcolor{resultDrop}{\ensuremath{(-6.97)}}} & 29.49\hspace*{7pt} \\
\modelentry{dolphin-v2}\nobreak\hspace{0.3em}{\fontsize{8.5}{9}\selectfont \href{https://github.com/bytedance/Dolphin}{\textcolor{black!65}{(2025-12)}}} & 65.90\hspace*{7pt} & \resulttrack{60.24}{\textcolor{resultDrop}{\ensuremath{(-5.66)}}} & \resulttrack{44.92}{\textcolor{resultDrop}{\ensuremath{(-20.98)}}} & 57.02\hspace*{7pt} \\
\rowcolor{resultStripe}
\modelentry{teleocr}\nobreak\hspace{0.3em}{\fontsize{8.5}{9}\selectfont \href{https://huggingface.co/StarDoc-AI/TeleOCR}{\textcolor{black!65}{(2026-08)}}} & \textbf{87.16}\hspace*{7pt} & \resulttrack{\underline{79.40}}{\textcolor{resultDrop}{\ensuremath{(-7.76)}}} & \resulttrack{69.08}{\textcolor{resultDrop}{\ensuremath{(-18.08)}}} & \underline{78.55}\hspace*{7pt} \\
\addlinespace[2pt]
\rowcolor{e2eBand}\multicolumn{5}{l}{\strut\textbf{End-to-end specialists}} \\
\rowcolor{resultStripe}
\modelentry{ovisocr2}\nobreak\hspace{0.3em}{\fontsize{8.5}{9}\selectfont \href{https://arxiv.org/abs/2607.13639}{\textcolor{black!65}{(2026-07)}}} & 81.53\hspace*{7pt} & \resulttrack{77.29}{\textcolor{resultDrop}{\ensuremath{(-4.24)}}} & \resulttrack{66.49}{\textcolor{resultDrop}{\ensuremath{(-15.04)}}} & 75.10\hspace*{7pt} \\
\modelentry{abot-ocr}\nobreak\hspace{0.3em}{\fontsize{8.5}{9}\selectfont \href{https://github.com/amap-cvlab/ABot-OCR}{\textcolor{black!65}{(2026-05)}}} & 76.92\hspace*{7pt} & \resulttrack{72.43}{\textcolor{resultDrop}{\ensuremath{(-4.49)}}} & \resulttrack{61.77}{\textcolor{resultDrop}{\ensuremath{(-15.15)}}} & 70.37\hspace*{7pt} \\
\rowcolor{resultStripe}
\modelentry{qianfan-ocr}\nobreak\hspace{0.3em}{\fontsize{8.5}{9}\selectfont \href{https://github.com/baidubce/Qianfan-VL}{\textcolor{black!65}{(2026-03)}}} & 57.22\hspace*{7pt} & \resulttrack{50.85}{\textcolor{resultDrop}{\ensuremath{(-6.37)}}} & \resulttrack{45.06}{\textcolor{resultDrop}{\ensuremath{(-12.16)}}} & 51.04\hspace*{7pt} \\
\modelentry{unlimited-ocr}\nobreak\hspace{0.3em}{\fontsize{8.5}{9}\selectfont \href{https://huggingface.co/baidu/Unlimited-OCR}{\textcolor{black!65}{(2026-06)}}} & 72.05\hspace*{7pt} & \resulttrack{65.30}{\textcolor{resultDrop}{\ensuremath{(-6.75)}}} & \resulttrack{53.71}{\textcolor{resultDrop}{\ensuremath{(-18.34)}}} & 63.69\hspace*{7pt} \\
\rowcolor{resultStripe}
\modelentry{wevisdoc-4b}\nobreak\hspace{0.3em}{\fontsize{8.5}{9}\selectfont \href{https://github.com/Tencent/WeVisDoc}{\textcolor{black!65}{(2026-09)}}} & 79.59\hspace*{7pt} & \resulttrack{77.99}{\textcolor{resultDrop}{\ensuremath{(-1.60)}}} & \resulttrack{69.14}{\textcolor{resultDrop}{\ensuremath{(-10.45)}}} & 75.57\hspace*{7pt} \\
\modelentry{wevisdoc-2b}\nobreak\hspace{0.3em}{\fontsize{8.5}{9}\selectfont \href{https://github.com/Tencent/WeVisDoc}{\textcolor{black!65}{(2026-09)}}} & 79.08\hspace*{7pt} & \resulttrack{76.33}{\textcolor{resultDrop}{\ensuremath{(-2.75)}}} & \resulttrack{65.68}{\textcolor{resultDrop}{\ensuremath{(-13.40)}}} & 73.69\hspace*{7pt} \\
\rowcolor{resultStripe}
\modelentry{hunyuanocr-1-5}\nobreak\hspace{0.3em}{\fontsize{8.5}{9}\selectfont \href{https://github.com/Tencent-Hunyuan/HunyuanOCR}{\textcolor{black!65}{(2026-07)}}} & 73.09\hspace*{7pt} & \resulttrack{69.36}{\textcolor{resultDrop}{\ensuremath{(-3.73)}}} & \resulttrack{60.28}{\textcolor{resultDrop}{\ensuremath{(-12.81)}}} & 67.58\hspace*{7pt} \\
\modelentry{hunyuanocr}\nobreak\hspace{0.3em}{\fontsize{8.5}{9}\selectfont \href{https://github.com/Tencent-Hunyuan/HunyuanOCR}{\textcolor{black!65}{(2025-11)}}} & 65.61\hspace*{7pt} & \resulttrack{61.49}{\textcolor{resultDrop}{\ensuremath{(-4.12)}}} & \resulttrack{54.58}{\textcolor{resultDrop}{\ensuremath{(-11.03)}}} & 60.56\hspace*{7pt} \\
\rowcolor{resultStripe}
\modelentry{dots-ocr}\nobreak\hspace{0.3em}{\fontsize{8.5}{9}\selectfont \href{https://huggingface.co/dots-studio/dots.ocr}{\textcolor{black!65}{(2025-07)}}} & 72.01\hspace*{7pt} & \resulttrack{65.95}{\textcolor{resultDrop}{\ensuremath{(-6.06)}}} & \resulttrack{55.68}{\textcolor{resultDrop}{\ensuremath{(-16.33)}}} & 64.55\hspace*{7pt} \\
\modelentry{firered-ocr}\nobreak\hspace{0.3em}{\fontsize{8.5}{9}\selectfont \href{https://huggingface.co/FireRedTeam/FireRed-OCR}{\textcolor{black!65}{(2026-02)}}} & 70.81\hspace*{7pt} & \resulttrack{68.49}{\textcolor{resultDrop}{\ensuremath{(-2.32)}}} & \resulttrack{57.42}{\textcolor{resultDrop}{\ensuremath{(-13.39)}}} & 65.57\hspace*{7pt} \\
\rowcolor{resultStripe}
\modelentry{deepseek-ocr-2}\nobreak\hspace{0.3em}{\fontsize{8.5}{9}\selectfont \href{https://arxiv.org/abs/2601.20552}{\textcolor{black!65}{(2026-01)}}} & 55.53\hspace*{7pt} & \resulttrack{49.41}{\textcolor{resultDrop}{\ensuremath{(-6.12)}}} & \resulttrack{43.60}{\textcolor{resultDrop}{\ensuremath{(-11.93)}}} & 49.51\hspace*{7pt} \\
\modelentry{deepseek-ocr}\nobreak\hspace{0.3em}{\fontsize{8.5}{9}\selectfont \href{https://huggingface.co/deepseek-ai/DeepSeek-OCR/commit/59512895521e669adb87064b37e5749fe9b9f5d2}{\textcolor{black!65}{(2025-10)}}} & 53.50\hspace*{7pt} & \resulttrack{46.95}{\textcolor{resultDrop}{\ensuremath{(-6.55)}}} & \resulttrack{40.48}{\textcolor{resultDrop}{\ensuremath{(-13.02)}}} & 46.98\hspace*{7pt} \\
\rowcolor{resultStripe}
\modelentry{olmocr-2-7b}\nobreak\hspace{0.3em}{\fontsize{8.5}{9}\selectfont \href{https://github.com/allenai/olmocr}{\textcolor{black!65}{(2025-10)}}} & 69.36\hspace*{7pt} & \resulttrack{65.87}{\textcolor{resultDrop}{\ensuremath{(-3.49)}}} & \resulttrack{56.10}{\textcolor{resultDrop}{\ensuremath{(-13.26)}}} & 63.78\hspace*{7pt} \\
\modelentry{olmocr-7b}\nobreak\hspace{0.3em}{\fontsize{8.5}{9}\selectfont \href{https://github.com/allenai/olmocr}{\textcolor{black!65}{(2025-08)}}} & 62.56\hspace*{7pt} & \resulttrack{57.84}{\textcolor{resultDrop}{\ensuremath{(-4.72)}}} & \resulttrack{47.30}{\textcolor{resultDrop}{\ensuremath{(-15.26)}}} & 55.90\hspace*{7pt} \\
\rowcolor{resultStripe}
\modelentry{ocrverse}\nobreak\hspace{0.3em}{\fontsize{8.5}{9}\selectfont \href{https://github.com/DocTron-hub/OCRVerse}{\textcolor{black!65}{(2026-01)}}} & 73.18\hspace*{7pt} & \resulttrack{71.36}{\textcolor{resultDrop}{\ensuremath{(-1.82)}}} & \resulttrack{63.66}{\textcolor{resultDrop}{\ensuremath{(-9.52)}}} & 69.40\hspace*{7pt} \\
\modelentry{fd-rl}\nobreak\hspace{0.3em}{\fontsize{8.5}{9}\selectfont \href{https://github.com/DocTron-hub/FD-RL}{\textcolor{black!65}{(2025-11)}}} & 78.38\hspace*{7pt} & \resulttrack{76.33}{\textcolor{resultDrop}{\ensuremath{(-2.05)}}} & \resulttrack{67.04}{\textcolor{resultDrop}{\ensuremath{(-11.34)}}} & 73.92\hspace*{7pt} \\
\rowcolor{resultStripe}
\modelentry{nanonets-ocr2}\nobreak\hspace{0.3em}{\fontsize{8.5}{9}\selectfont \href{https://benchmarks.nanonets.com/}{\textcolor{black!65}{(2025-10)}}} & 64.83\hspace*{7pt} & \resulttrack{61.23}{\textcolor{resultDrop}{\ensuremath{(-3.60)}}} & \resulttrack{49.03}{\textcolor{resultDrop}{\ensuremath{(-15.80)}}} & 58.36\hspace*{7pt} \\
\modelentry{ocrflux-3b}\nobreak\hspace{0.3em}{\fontsize{8.5}{9}\selectfont \href{https://github.com/chatdoc-com/OCRFlux}{\textcolor{black!65}{(2025-06)}}} & 47.14\hspace*{7pt} & \resulttrack{41.82}{\textcolor{resultDrop}{\ensuremath{(-5.32)}}} & \resulttrack{37.21}{\textcolor{resultDrop}{\ensuremath{(-9.93)}}} & 42.06\hspace*{7pt} \\
\rowcolor{resultStripe}
\modelentry{logics-parsing-v2}\nobreak\hspace{0.3em}{\fontsize{8.5}{9}\selectfont \href{https://huggingface.co/Logics-MLLM/Logics-Parsing-v2}{\textcolor{black!65}{(2026-02)}}} & 76.35\hspace*{7pt} & \resulttrack{73.85}{\textcolor{resultDrop}{\ensuremath{(-2.50)}}} & \resulttrack{67.64}{\textcolor{resultDrop}{\ensuremath{(-8.71)}}} & 72.61\hspace*{7pt} \\
\addlinespace[2pt]
\rowcolor{vlmBand}\multicolumn{5}{l}{\strut\textbf{General-purpose VLMs}} \\
\rowcolor{resultStripe}
\modelentry{qwen3-8-27b}\nobreak\hspace{0.3em}{\fontsize{8.5}{9}\selectfont \href{https://github.com/QwenLM/Qwen3.8#news}{\textcolor{black!65}{(2026-08)}}} & 80.45\hspace*{7pt} & \resulttrack{78.86}{\textcolor{resultDrop}{\ensuremath{(-1.59)}}} & \resulttrack{73.59}{\textcolor{resultDrop}{\ensuremath{(-6.86)}}} & 77.63\hspace*{7pt} \\
\modelentry{qwen3-8-flash-next}\nobreak\hspace{0.3em}{\fontsize{8.5}{9}\selectfont \href{https://github.com/QwenLM/Qwen3.8-Flash-Next#news}{\textcolor{black!65}{(2026-08)}}} & 72.23\hspace*{7pt} & \resulttrack{70.96}{\textcolor{resultDrop}{\ensuremath{(-1.27)}}} & \resulttrack{62.42}{\textcolor{resultDrop}{\ensuremath{(-9.81)}}} & 68.54\hspace*{7pt} \\
\rowcolor{resultStripe}
\modelentry{qwen3-6-35b-a3b}\nobreak\hspace{0.3em}{\fontsize{8.5}{9}\selectfont \href{https://github.com/QwenLM/Qwen3.8#news}{\textcolor{black!65}{(2026-04)}}} & 72.14\hspace*{7pt} & \resulttrack{69.16}{\textcolor{resultDrop}{\ensuremath{(-2.98)}}} & \resulttrack{60.12}{\textcolor{resultDrop}{\ensuremath{(-12.02)}}} & 67.14\hspace*{7pt} \\
\modelentry{qwen3-6-27b}\nobreak\hspace{0.3em}{\fontsize{8.5}{9}\selectfont \href{https://github.com/QwenLM/Qwen3.8#news}{\textcolor{black!65}{(2026-04)}}} & 70.20\hspace*{7pt} & \resulttrack{67.49}{\textcolor{resultDrop}{\ensuremath{(-2.71)}}} & \resulttrack{59.18}{\textcolor{resultDrop}{\ensuremath{(-11.02)}}} & 65.62\hspace*{7pt} \\
\rowcolor{resultStripe}
\modelentry{qwen3-5-397b-a17b}\nobreak\hspace{0.3em}{\fontsize{8.5}{9}\selectfont \href{https://github.com/QwenLM/Qwen3.8#news}{\textcolor{black!65}{(2026-02)}}} & 69.12\hspace*{7pt} & \resulttrack{68.34}{\textcolor{resultDrop}{\ensuremath{(-0.78)}}} & \resulttrack{62.70}{\textcolor{resultDrop}{\ensuremath{(-6.42)}}} & 66.72\hspace*{7pt} \\
\modelentry{qwen3-5-122b-a10b}\nobreak\hspace{0.3em}{\fontsize{8.5}{9}\selectfont \href{https://github.com/QwenLM/Qwen3.8#news}{\textcolor{black!65}{(2026-02)}}} & 76.14\hspace*{7pt} & \resulttrack{76.34}{\textcolor{resultGain}{\ensuremath{(+0.20)}}} & \resulttrack{69.85}{\textcolor{resultDrop}{\ensuremath{(-6.29)}}} & 74.11\hspace*{7pt} \\
\rowcolor{resultStripe}
\modelentry{qwen3-5-35b-a3b}\nobreak\hspace{0.3em}{\fontsize{8.5}{9}\selectfont \href{https://github.com/QwenLM/Qwen3.8#news}{\textcolor{black!65}{(2026-02)}}} & 68.40\hspace*{7pt} & \resulttrack{68.04}{\textcolor{resultDrop}{\ensuremath{(-0.36)}}} & \resulttrack{60.59}{\textcolor{resultDrop}{\ensuremath{(-7.81)}}} & 65.68\hspace*{7pt} \\
\modelentry{qwen3-5-27b}\nobreak\hspace{0.3em}{\fontsize{8.5}{9}\selectfont \href{https://github.com/QwenLM/Qwen3.8#news}{\textcolor{black!65}{(2026-02)}}} & 72.07\hspace*{7pt} & \resulttrack{70.73}{\textcolor{resultDrop}{\ensuremath{(-1.34)}}} & \resulttrack{65.92}{\textcolor{resultDrop}{\ensuremath{(-6.15)}}} & 69.57\hspace*{7pt} \\
\rowcolor{resultStripe}
\modelentry{qwen3-5-9b}\nobreak\hspace{0.3em}{\fontsize{8.5}{9}\selectfont \href{https://github.com/QwenLM/Qwen3.8#news}{\textcolor{black!65}{(2026-03)}}} & 73.87\hspace*{7pt} & \resulttrack{73.34}{\textcolor{resultDrop}{\ensuremath{(-0.53)}}} & \resulttrack{65.45}{\textcolor{resultDrop}{\ensuremath{(-8.42)}}} & 70.89\hspace*{7pt} \\
\modelentry{qwen3-5-4b}\nobreak\hspace{0.3em}{\fontsize{8.5}{9}\selectfont \href{https://github.com/QwenLM/Qwen3.8#news}{\textcolor{black!65}{(2026-03)}}} & 73.45\hspace*{7pt} & \resulttrack{72.53}{\textcolor{resultDrop}{\ensuremath{(-0.92)}}} & \resulttrack{63.47}{\textcolor{resultDrop}{\ensuremath{(-9.98)}}} & 69.82\hspace*{7pt} \\
\rowcolor{resultStripe}
\modelentry{qwen3-5-2b}\nobreak\hspace{0.3em}{\fontsize{8.5}{9}\selectfont \href{https://github.com/QwenLM/Qwen3.8#news}{\textcolor{black!65}{(2026-03)}}} & 66.24\hspace*{7pt} & \resulttrack{65.22}{\textcolor{resultDrop}{\ensuremath{(-1.02)}}} & \resulttrack{55.92}{\textcolor{resultDrop}{\ensuremath{(-10.32)}}} & 62.46\hspace*{7pt} \\
\modelentry{qwen3-5-0-8b}\nobreak\hspace{0.3em}{\fontsize{8.5}{9}\selectfont \href{https://github.com/QwenLM/Qwen3.8#news}{\textcolor{black!65}{(2026-03)}}} & 60.77\hspace*{7pt} & \resulttrack{59.28}{\textcolor{resultDrop}{\ensuremath{(-1.49)}}} & \resulttrack{47.93}{\textcolor{resultDrop}{\ensuremath{(-12.84)}}} & 55.99\hspace*{7pt} \\
\rowcolor{resultStripe}
\modelentry{qwen3-vl-8b}\nobreak\hspace{0.3em}{\fontsize{8.5}{9}\selectfont \href{https://github.com/QwenLM/Qwen3-VL#news}{\textcolor{black!65}{(2025-10)}}} & 72.44\hspace*{7pt} & \resulttrack{72.03}{\textcolor{resultDrop}{\ensuremath{(-0.41)}}} & \resulttrack{62.73}{\textcolor{resultDrop}{\ensuremath{(-9.71)}}} & 69.07\hspace*{7pt} \\
\modelentry{qwen3-vl-4b}\nobreak\hspace{0.3em}{\fontsize{8.5}{9}\selectfont \href{https://github.com/QwenLM/Qwen3-VL#news}{\textcolor{black!65}{(2025-10)}}} & 72.04\hspace*{7pt} & \resulttrack{70.84}{\textcolor{resultDrop}{\ensuremath{(-1.20)}}} & \resulttrack{59.61}{\textcolor{resultDrop}{\ensuremath{(-12.43)}}} & 67.50\hspace*{7pt} \\
\rowcolor{resultStripe}
\modelentry{qwen3-vl-2b}\nobreak\hspace{0.3em}{\fontsize{8.5}{9}\selectfont \href{https://github.com/QwenLM/Qwen3-VL#news}{\textcolor{black!65}{(2025-10)}}} & 66.37\hspace*{7pt} & \resulttrack{65.81}{\textcolor{resultDrop}{\ensuremath{(-0.56)}}} & \resulttrack{54.09}{\textcolor{resultDrop}{\ensuremath{(-12.28)}}} & 62.09\hspace*{7pt} \\
\modelentry{deepseek-v4-flash-vision-exp}\nobreak\hspace{0.3em}{\fontsize{8.5}{9}\selectfont \href{https://api-docs.deepseek.com/updates/}{\textcolor{black!65}{(2026-08)}}} & 54.69\hspace*{7pt} & \resulttrack{52.98}{\textcolor{resultDrop}{\ensuremath{(-1.71)}}} & \resulttrack{42.68}{\textcolor{resultDrop}{\ensuremath{(-12.01)}}} & 50.12\hspace*{7pt} \\
\rowcolor{resultStripe}
\modelentry{glm-5-3-flash}\nobreak\hspace{0.3em}{\fontsize{8.5}{9}\selectfont \href{https://docs.z.ai/release-notes/new-released}{\textcolor{black!65}{(2026-08)}}} & \underline{83.04}\hspace*{7pt} & \resulttrack{\textbf{81.03}}{\textcolor{resultDrop}{\ensuremath{(-2.01)}}} & \resulttrack{74.99}{\textcolor{resultDrop}{\ensuremath{(-8.05)}}} & \textbf{79.69}\hspace*{7pt} \\
\modelentry{minicpm-v-4-6}\nobreak\hspace{0.3em}{\fontsize{8.5}{9}\selectfont \href{https://github.com/OpenBMB/MiniCPM-V#news}{\textcolor{black!65}{(2026-05)}}} & 64.47\hspace*{7pt} & \resulttrack{59.04}{\textcolor{resultDrop}{\ensuremath{(-5.43)}}} & \resulttrack{48.01}{\textcolor{resultDrop}{\ensuremath{(-16.46)}}} & 57.17\hspace*{7pt} \\
\rowcolor{resultStripe}
\modelentry{minicpm-v-4-5}\nobreak\hspace{0.3em}{\fontsize{8.5}{9}\selectfont \href{https://github.com/OpenBMB/MiniCPM-V#news}{\textcolor{black!65}{(2025-08)}}} & 51.81\hspace*{7pt} & \resulttrack{49.38}{\textcolor{resultDrop}{\ensuremath{(-2.43)}}} & \resulttrack{37.59}{\textcolor{resultDrop}{\ensuremath{(-14.22)}}} & 46.26\hspace*{7pt} \\
\modelentry{gemini-3-6-flash}\nobreak\hspace{0.3em}{\fontsize{8.5}{9}\selectfont \href{https://ai.google.dev/gemini-api/docs/changelog#july-21-2026}{\textcolor{black!65}{(2026-07)}}} & 79.97\hspace*{7pt} & \resulttrack{79.11}{\textcolor{resultDrop}{\ensuremath{(-0.86)}}} & \resulttrack{\textbf{75.73}}{\textcolor{resultDrop}{\ensuremath{(-4.24)}}} & 78.27\hspace*{7pt} \\
\rowcolor{resultStripe}
\modelentry{gemini-3-1-pro}\nobreak\hspace{0.3em}{\fontsize{8.5}{9}\selectfont \href{https://ai.google.dev/gemini-api/docs/changelog#february-19-2026}{\textcolor{black!65}{(2026-02)}}} & 70.04\hspace*{7pt} & \resulttrack{69.28}{\textcolor{resultDrop}{\ensuremath{(-0.76)}}} & \resulttrack{71.98}{\textcolor{resultGain}{\ensuremath{(+1.94)}}} & 70.43\hspace*{7pt} \\
\modelentry{claude-opus-5}\nobreak\hspace{0.3em}{\fontsize{8.5}{9}\selectfont \href{https://www.anthropic.com/news/claude-opus-5}{\textcolor{black!65}{(2026-07)}}} & 82.99\hspace*{7pt} & \resulttrack{72.94}{\textcolor{resultDrop}{\ensuremath{(-10.05)}}} & \resulttrack{\underline{75.27}}{\textcolor{resultDrop}{\ensuremath{(-7.72)}}} & 77.07\hspace*{7pt} \\
\rowcolor{resultStripe}
\modelentry{gpt-5-6-sol}\nobreak\hspace{0.3em}{\fontsize{8.5}{9}\selectfont \href{https://openai.com/index/gpt-5-6/}{\textcolor{black!65}{(2026-07)}}} & 74.15\hspace*{7pt} & \resulttrack{71.73}{\textcolor{resultDrop}{\ensuremath{(-2.42)}}} & \resulttrack{61.59}{\textcolor{resultDrop}{\ensuremath{(-12.56)}}} & 69.16\hspace*{7pt} \\
\modelentry{minimax-m3}\nobreak\hspace{0.3em}{\fontsize{8.5}{9}\selectfont \href{https://www.minimax.io/blog/minimax-m3}{\textcolor{black!65}{(2026-06)}}} & 55.78\hspace*{7pt} & \resulttrack{53.85}{\textcolor{resultDrop}{\ensuremath{(-1.93)}}} & \resulttrack{42.54}{\textcolor{resultDrop}{\ensuremath{(-13.24)}}} & 50.72\hspace*{7pt} \\
\rowcolor{resultStripe}
\modelentry{kimi-k2-6}\nobreak\hspace{0.3em}{\fontsize{8.5}{9}\selectfont \href{https://www.kimi.com/en/help/agent/agent-overview}{\textcolor{black!65}{(2026-04)}}} & 72.32\hspace*{7pt} & \resulttrack{69.95}{\textcolor{resultDrop}{\ensuremath{(-2.37)}}} & \resulttrack{68.02}{\textcolor{resultDrop}{\ensuremath{(-4.30)}}} & 70.10\hspace*{7pt} \\
\modelentry{step3-vl}\nobreak\hspace{0.3em}{\fontsize{8.5}{9}\selectfont \href{https://github.com/stepfun-ai/Step3-VL-10B/actions/runs/21027808364}{\textcolor{black!65}{(2026-01)}}} & 53.65\hspace*{7pt} & \resulttrack{52.74}{\textcolor{resultDrop}{\ensuremath{(-0.91)}}} & \resulttrack{45.06}{\textcolor{resultDrop}{\ensuremath{(-8.59)}}} & 50.48\hspace*{7pt} \\
\bottomrule
\end{tabular}
}
\end{table}

\subsection{Component Analysis}
\label{sec:diagnosis}

\begin{figure}[!ht]
\centering
\input{data/component_heatmap_links.tex}
\caption{\textbf{Component profiles of 58 models.} Three-track means reveal distinct strengths in text, formulas, tables, and reading order. Text and Reading use $100(1{-}\mathrm{Edit})$, Formula uses CDM, and Table uses TEDS. All scores range from 0 to 100, with higher values indicating better performance.}
\label{fig:component_heatmap}
\end{figure}

\paragraph{Component strengths vary across models and architectures.} In the three-track averages of Figure~\ref{fig:component_heatmap}, Qwen3.8-27B leads Text (78.00), Gemini~3.6~Flash leads Formula (77.81), GLM-5.3-Flash leads Table (87.35), and TeleOCR leads Reading (75.14). Averaging equally over models within each architecture, E2E specialists achieve the highest mean Text/Reading scores (66.18/57.44), while general VLMs lead Formula/Table (62.48/70.07). The general-VLM advantage over E2E specialists is larger for Formula (7.39 points) than for Table (1.83 points). At the model level, the Avg$_3$ leader GLM-5.3-Flash scores 60.37 on Reading, 14.77 points below TeleOCR. Reading order is evaluated separately from Overall and measures whether content is recovered in the correct sequence, including transitions between columns, text blocks, and other page elements.

\Needspace{74pt}
\subsection{Case Study}
\label{sec:casestudy}

\begin{figure}[!t]
\centering
\includegraphics[width=\textwidth]{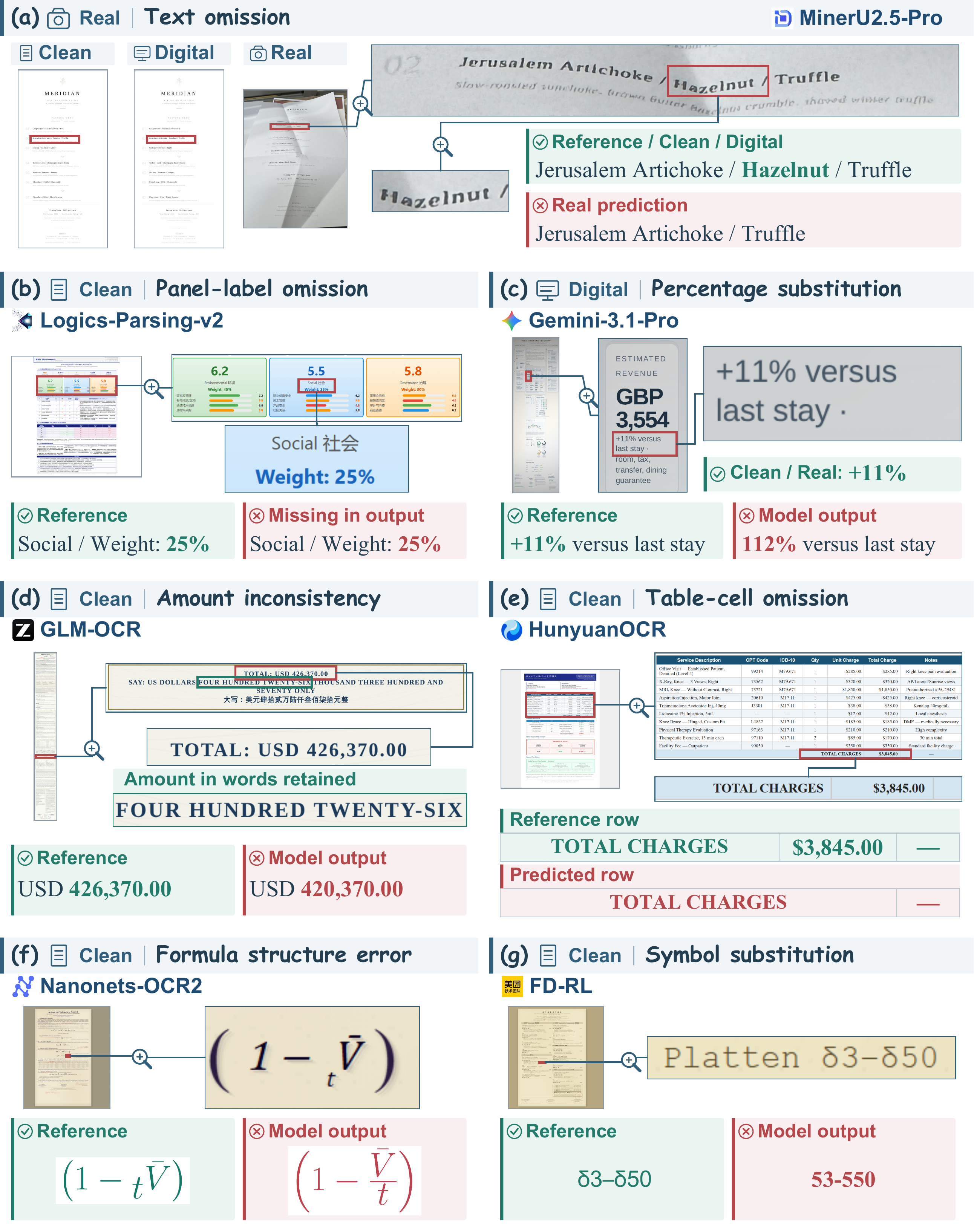}
\caption{\textbf{Parsing failures across three image tracks.} Examples include (a) omitted ingredients, (b) missing panel labels, (c) altered percentages, (d) inconsistent numeric and written amounts, (e) a missing table cell, (f) a subscript rewritten as division, and (g) Greek symbols read as digits. These errors affect content completeness, numerical accuracy, and document structure.}
\label{fig:case_study_main}
\end{figure}

Figure~\ref{fig:case_study_main} shows omissions, numerical errors, and failures in table structure, formulas, and symbol recognition across the three tracks. Appendix~\ref{app:examples} provides 20 additional cases on different document pages.

\Needspace{4\baselineskip}
\paragraph{Clean: omissions, values, and symbols.} In Figure~\ref{fig:case_study_main}(b), Logics-Parsing-v2 omits the Social label and its 25\% weight but retains the neighboring labels and all three ESG scores. In Figure~\ref{fig:case_study_main}(d), GLM-OCR changes a bill-of-lading total from USD~426,370.00 to USD~420,370.00 while preserving the correct amount in words.

In Figure~\ref{fig:case_study_main}(e), HunyuanOCR omits the \$3,845.00 amount cell from the medical bill's \emph{Itemized Charges} total row and expands the preceding label cell from five to six columns. The same amount remains in the following table, localizing the omission to the first table's total row. Nanonets-OCR2 converts the time subscript of ${}_{t}\bar{V}$ into division by $t$ in an actuarial formula (Figure~\ref{fig:case_study_main}(f)). FD-RL reads the thickness notation ``$\delta$3--$\delta$50'' as ``53-550'' (Figure~\ref{fig:case_study_main}(g)).

\paragraph{Digital: percentage substitution.} In a hotel reservation dashboard with uneven lighting, Gemini-3.1-Pro changes a clearly readable revenue-growth percentage from +11\% to 112\% (Figure~\ref{fig:case_study_main}(c)). Its paired Clean and Real predictions preserve +11\%.

\paragraph{Real: omitted content.} On an oblique photograph of a creased menu, MinerU2.5-Pro omits ``Hazelnut'' from ``Jerusalem Artichoke / Hazelnut / Truffle'' (Figure~\ref{fig:case_study_main}(a)). Its corresponding Clean and Digital predictions preserve the complete ingredient list.

\section{Conclusion and Future Work}
\label{sec:conclusion}

\textbf{Conclusion.} \ours{} provides source-linked annotations and matched Clean, Digital, and Real images across 10 domains and 66 subcategories. Results from 58 models show room for improvement across parsing components and image conditions. Compact specialists can match or outperform much larger general models. We release the benchmark, document sources, generation tools, and evaluation code to support reproducible comparisons and further research.

\Needspace{7\baselineskip}
\textbf{Limitations and Future Work.} (i) We adopt OmniDocBench's evaluation framework, whose metrics need refinement to better capture semantic omissions and structural consistency. (ii) HTML rendering has limited coverage of handwriting and historical books; these settings require additional real-document data. (iii) Human review identified 83 Real images that are very difficult to read, mainly recaptures of long pages. Appendix~\ref{app:readability} analyzes these cases separately: excluding them leaves model rankings largely stable and preserves our main comparative conclusions. Future acquisition should retain readable detail and record quality flags to separate capture limits from model errors.

\paragraph{AI Use Statement.} AI tools were used for manuscript formatting, data organization, and discussion of research ideas.

\paragraph{Ethics Statement.} \ours{} uses generated documents and recaptures of those pages for research evaluation. Applications involving sensitive information should include privacy safeguards and human verification.

\paragraph{Reproducibility Statement.} The code link is provided in the abstract.

\bibliographystyle{iclr2027_conference}
\bibliography{references}

\newpage
\appendix

\section*{Appendix Contents}
\begingroup
\setlength{\parskip}{3pt}
\newcommand{\appendixcontentsentry}[1]{%
  \noindent\hyperref[#1]{%
    \makebox[1.6em][l]{\ref*{#1}}\nameref*{#1}\dotfill\pageref*{#1}}\par}
\appendixcontentsentry{app:benchmark_comparison}
\appendixcontentsentry{app:examples}
\appendixcontentsentry{app:readability}
\appendixcontentsentry{app:full_results}
\appendixcontentsentry{app:annotation_errors}
\appendixcontentsentry{app:degradation_params}
\appendixcontentsentry{app:categories}
\endgroup
\vspace{6pt}

\section{Benchmark Comparison: Scope, Versions, and Evidence}
\label{app:benchmark_comparison}

Table~\ref{tab:benchmark_comparison} summarizes complementary datasets using the papers and author-maintained releases reviewed on September 25, 2026. Its grouping separates component recognition, document understanding and synthetic supervision, and full-document parsing. The layout is inspired by the task-oriented comparison in MADQA~\citep{madqa}; the dimensions here are selected for parsing and annotation provenance.

\paragraph{Definitions.} \emph{Corpus size} retains the units of the cited publication or named release. Historical training-and-test corpora and evaluation-only sets are both included; their sizes are not directly comparable measures of evaluation strength. Rounded ``k'' counts indicate thousands. The four \emph{parsing target} columns cover text transcription, table structure, formula recognition, and reading order. Layout detection and question answering provide complementary context outside the displayed task columns. A check denotes an explicitly evaluated task; a half-filled circle denotes a restricted subtask or subset; a dash denotes a dimension not reported in the reviewed protocol. These symbols summarize scope, not metric quality. QA answers and auxiliary OCR strings alone do not count as a text-transcription task. The \emph{label source} icons denote markup-derived targets, model assistance, human annotation/review, and inherited labels; combinations summarize contributing processes. Markup derivation does not imply that every resource retains regenerable sources. The \emph{paired image variants} columns distinguish digital transformations from physical acquisitions of the same underlying page. Naturally noisy images alone do not establish pairing. Multipliers denote base pages times views: Real5 counts five physical captures; MDPBench and PureDocBench include the original in their four and three views, respectively.

\paragraph{Restricted task scopes.} PubTabNet's text task covers table cells. OCRBench's formula subset contains 100 handwritten expressions. olmOCR-bench explicitly tests content, formulas, and reading order; its table tests concern cell relations. Its markup icon refers to the arXiv-math subset. ParseBench evaluates text, tables, reading order, and layout grounding; its formula check concerns \LaTeX{} notation rather than complete formula-content attribution. MPDocBench-Parse additionally evaluates figure extraction, hierarchy, and cross-page continuity. DocPTBench's physical captures cover a 100-page subset of its 981 base pages. A missing label-source entry for LogicsDocBench records undisclosed benchmark-specific construction details. Additional capabilities, including layout, question answering, hierarchy, translation, charts, code, and catalogs, are detailed below.

\paragraph{Component datasets.} PubLayNet~\citep{publaynet} reports over 360,000 document images with PDF--XML alignment, while DocBank~\citep{docbank} uses \LaTeX{} to label tokens on 500,000 pages. DocLayNet~\citep{doclaynet} provides 80,863 human-annotated pages for layout segmentation. TableBank~\citep{tablebank} contains 417,234 table instances on 278,582 page images for detection; its separate structure-recognition task contains 145,463 table images and predicts structural tags. PubTabNet~\citep{teds} pairs approximately 568,000 table crops with HTML and introduces TEDS. PubTables-1M~\citep{pubtables1m} contains 947,642 tables on 575,305 pages; canonicalization and functional labels enrich its structural evaluation. Its original detection-based recognition model uses separately supplied text for text-bearing outputs. The im2latex corpus~\citep{im2latex} contains 103,556 rendered expressions, rounded to 104k in the table. These resources establish clear precedents for source-derived supervision. ReadingBank~\citep{layoutreader} extends this lineage to Word-derived reading order, and HRDoc~\citep{hrdoc} targets document hierarchies. Form and receipt datasets~\citep{funsd,sroie,cord} and financial tables~\citep{fintabnet} provide further task/domain coverage.

\paragraph{Understanding tasks.} DocVQA~\citep{docvqa} uses 12,767 document images for question answering. OCRBench v2~\citep{ocrbenchv2} has 10,000 public instruction--answer samples and includes full-page OCR, element parsing, and reasoning; its table entry refers to this public set. Construction reuses existing dataset GT, adds human labels where needed, and checks instructions and answers. MADQA~\citep{madqa} uses 800 PDFs and 2,250 human-authored questions for retrieval and agentic answering. Other benchmarks discussed in Section~\ref{sec:comparison} test infographic reasoning~\citep{infographicvqa}, heterogeneous multi-page QA~\citep{dude}, long-document QA~\citep{mmlongbenchdoc}, or visual text reasoning and grounded perception~\citep{ocrbench,ocrreasoningbench,mmdocbench}. Answer correctness and evidence grounding complement the content/structure recovery measured here; our parsing scores do not establish downstream QA or agent performance.

\paragraph{OCR and grounded understanding.} The original OCRBench~\citep{ocrbench} contains 1,000 manually checked QA pairs, including 300 text-recognition and 100 handwritten-formula samples drawn from existing datasets. CC-OCR~\citep{ccocr} contains 7,058 images across 39 subsets, covering scene and multilingual OCR, document/table/formula parsing, and information extraction. Its labels combine inherited annotations, model preannotations, and manual correction; reading order has no separate evaluation metric. DOGR-Bench~\citep{dogr} contains 3,600 samples across seven grounding and referring tasks, built with rendered-source/PDF annotations and model-generated instructions. Its grounded-answer metrics assess text spans and their boxes; its four parsing columns therefore remain unreported under our task definitions. The larger DOGR training corpus is a separate resource. Physical captures in CC-OCR do not establish a paired acquisition protocol.

\paragraph{CC-OCR v2.} The August 2026 revision of CC-OCR v2~\citep{ccocrv2} reports 7,093 evaluation samples across five tracks and 16 subtasks, with ten document categories, 32 recognition languages, and ten diagnostic factors. It reconstructs CC-OCR through sample refinement, new collection, inherited labels, model-assisted quality control, and human annotation/review. Text recognition uses multi-set F1; document/formula parsing uses edit similarity and tables use TEDS. Reading order is discussed within parsing but has no separately reported score. Acquisition-factor labels characterize observed conditions; the protocol does not report paired digital/physical views of identical content.

\paragraph{Synthetic supervision and controlled evaluation.} CoSyn-400K~\citep{cosyn} is a \textbf{training corpus}: 400,000 code-rendered images with 2.7 million instruction-tuning examples. Its source-code-conditioned generation supplies related prior art for synthetic supervision; task scores are evaluated on separate benchmarks. SynthDocBench~\citep{synthdocbench} evaluates 1,788 questions over 200 synthetic reports, averaging 51.1 pages, using controlled HTML layouts and source-backed chart data. Labels combine model generation, metadata-based numerical checks, consistency filtering, and a 100-question human review. It evaluates long-document chart reading, cross-modal reasoning, and multi-hop QA. Its optional scanned-PDF renderer and resolution ablations do not define a matched digital/physical degradation suite, so paired-variant checks are left unreported.

\paragraph{READoc and OmniDocBench.} We use the ACL 2025 version of READoc~\citep{readoc}, with 3,576 documents from arXiv, GitHub, and Zenodo. Its pipeline derives Markdown targets from \LaTeX{}, Markdown, DOCX, and HTML, including source-to-PDF rendering. Together with Nougat~\citep{nougat}, it is relevant prior art for markup-aligned parsing. For OmniDocBench~\citep{omnidocbench}, the table uses \textbf{v1.5 / 1,355 pages}. Illustrative annotation errors from this version appear in Appendix~\ref{app:annotation_errors}.

\paragraph{Matched robustness datasets.} DocPTBench~\citep{docptbench} starts with 981 OmniDocBench pages, adds 981 digitally simulated photographs, and physically captures a 100-page subset under four conditions (400 images). Thus 1,381 counts altered images, while 2,362 counts originals plus altered images; the additionally released rectified derivatives are a separate set. Its parsing targets inherit the base annotations, and translations undergo checking. Real5-OmniDocBench~\citep{real5omnidocbench} applies five physical conditions to all 1,355 v1.5 pages, yielding 6,775 variants with inherited GT. MDPBench~\citep{mdpbench} contains 850 original pages in 17 languages and three photographs per page: 3,400 images including originals. These are concrete precedents for paired acquisition evaluation. PureDocBench pairs all 1,475 freshly generated pages across clean, digital, and real tracks and retains their common HTML/CSS sources for inspection and regeneration.

\paragraph{Recent parsing suites.} olmOCR-bench is introduced in the June 2025 revision of olmOCR~\citep{olmocr}; the 1,403-page / 7,010-test count follows Appendix F, Table 10. Its unit tests cover content, mathematical expressions, table relations, and order; targets combine model generation, human design, and review. The arXiv-math subset aligns to original TeX, with manual verification, as described in the \href{https://github.com/allenai/olmocr/blob/main/olmocr/bench/README.md}{benchmark documentation}. The 900-page document module in the Logics-Parsing-Omni report~\citep{omniparsingbench} is named \emph{LogicsDocBench}; OmniParsingBench denotes the broader suite. Its document evaluation reports text edit distance, formula CDM, table TEDS/TEDS-S, and reading order; its training-data annotation description does not by itself establish the benchmark GT procedure. ParseBench~\citep{parsebench} v3 reports 2,078 pages and 169,011 rules across tables, charts, content, semantic formatting, and grounding, with VLM-assisted targets and human verification.

\paragraph{Languages, multiple pages, and specialized content.} MORE~\citep{morebench} reports 149 languages and six tasks. Its \href{https://github.com/zimoqingfeng/MORE}{official repository} flags a count discrepancy between 1,237 pages in the collection description and 1,288 images in the statistics table; we use the transparent approximation $\sim$1.3k. MPDocBench-Parse~\citep{mpdocbenchparse} contains 433 documents / 3,246 pages, with model preannotations, human correction, and structural checks, and evaluates hierarchy and cross-page continuity alongside content recovery. Dr.~DocBench~\citep{drdocbench} contains 4,514 annotated pages across 52 subjects, selected using parser disagreement and corrected by humans with expert review for specialized content. FinixDocBench~\citep{finixdoc}, discussed in the text, covers approximately 5,000 pages across digital, photographed, large-page, and internal financial-workflow subsets; its report releases a subset, and paired original/photo identities are not specified. These settings complement our predominantly Chinese--English, page-level, source-rendered evaluation.

\clearpage
\section{Additional Case-Study Evidence}
\label{app:examples}
\begin{figure}[H]
\centering
\input{data/case_appendix_01_links.tex}
\caption{\textbf{Clean-track omissions and association errors.} A laboratory sidebar is omitted, and a skill score is copied from its neighboring label, affecting content completeness and field accuracy.}
\label{fig:case_appendix_01}
\label{fig:case_study_long}
\end{figure}

\clearpage
\begin{figure}[H]
\centering
\input{data/case_appendix_02_links.tex}
\caption{\textbf{Clean-track symbol and code errors.} A percent sign becomes a digit in a financial formula, and a code identifier loses its separator, altering a numerical value and a callable identifier.}
\label{fig:case_appendix_02}
\end{figure}

\clearpage
\begin{figure}[H]
\centering
\input{data/case_appendix_03_links.tex}
\caption{\textbf{Clean-track numbering and value errors.} A legal paragraph number and a business-plan percentage are altered, corrupting document references and a stated revenue assumption.}
\label{fig:case_appendix_03}
\end{figure}

\clearpage
\begin{figure}[H]
\centering
\input{data/case_appendix_04_links.tex}
\caption{\textbf{Clean-track numerical errors.} A pressure value shifts by six orders of magnitude, and an extra digit increases a timeout tenfold, altering scientific specifications and software settings.}
\label{fig:case_appendix_04}
\end{figure}

\clearpage
\begin{figure}[H]
\centering
\input{data/case_appendix_05_links.tex}
\caption{\textbf{Digital-track association and identifier errors.} Payment percentages are misassigned and an API event name changes; both fields are recovered on the corresponding Clean pages.}
\label{fig:case_appendix_05}
\end{figure}

\clearpage
\begin{figure}[H]
\centering
\input{data/case_appendix_06_links.tex}
\caption{\textbf{Digital-track label and omission errors.} An answer choice is relabeled and a registrar filing statement is omitted; paired Clean outputs preserve both pieces of information.}
\label{fig:case_appendix_06}
\end{figure}

\clearpage
\begin{figure}[H]
\centering
\input{data/case_appendix_07_links.tex}
\caption{\textbf{Digital-track date and completeness errors.} A contract's effective year changes and a prediction ends before the next section; the paired Clean outputs preserve both details.}
\label{fig:case_appendix_07}
\end{figure}

\clearpage
\begin{figure}[H]
\centering
\input{data/case_appendix_08_links.tex}
\caption{\textbf{Real-track mathematical and numerical errors.} A formula gains a minus sign and a headline percentage changes, altering an exponential factor and a quantitative news claim.}
\label{fig:case_appendix_08}
\end{figure}

\clearpage
\begin{figure}[H]
\centering
\input{data/case_appendix_09_links.tex}
\caption{\textbf{Real-track heading and range errors.} A poster heading loses part of a method name, while altered endpoints of a subscription-price range misstate the service's pricing.}
\label{fig:case_appendix_09}
\end{figure}

\clearpage
\begin{figure}[H]
\centering
\input{data/case_appendix_10_links.tex}
\caption{\textbf{Real-track omission and duplication errors.} An experience count is omitted, and an address repeats with an incorrect postcode, corrupting document completeness and structure.}
\label{fig:case_appendix_10}
\end{figure}

\clearpage
\newcommand{\ReadabilityCohortSize}{51}
\newcommand{\ReadabilityExcludedPages}{83}
\newcommand{\ReadabilityRetainedRealPages}{1392}
\newcommand{\ReadabilityRealMeanGain}{2.32}
\newcommand{\ReadabilityRealMedianGain}{2.34}
\newcommand{\ReadabilityRealSpearman}{0.9983}
\newcommand{\ReadabilityRealMaxRankShift}{2}
\newcommand{\ReadabilityRealTopTenOverlap}{10}
\newcommand{\ReadabilityAvgMeanGain}{0.77}
\newcommand{\ReadabilityAvgMedianGain}{0.78}
\newcommand{\ReadabilityAvgSpearman}{0.9991}
\newcommand{\ReadabilityAvgMaxRankShift}{2}
\newcommand{\ReadabilityAvgTopTenOverlap}{10}

\section{Real-Track Readability Audit}
\label{app:readability}

\begin{figure}[H]
\centering
\includegraphics[width=\textwidth]{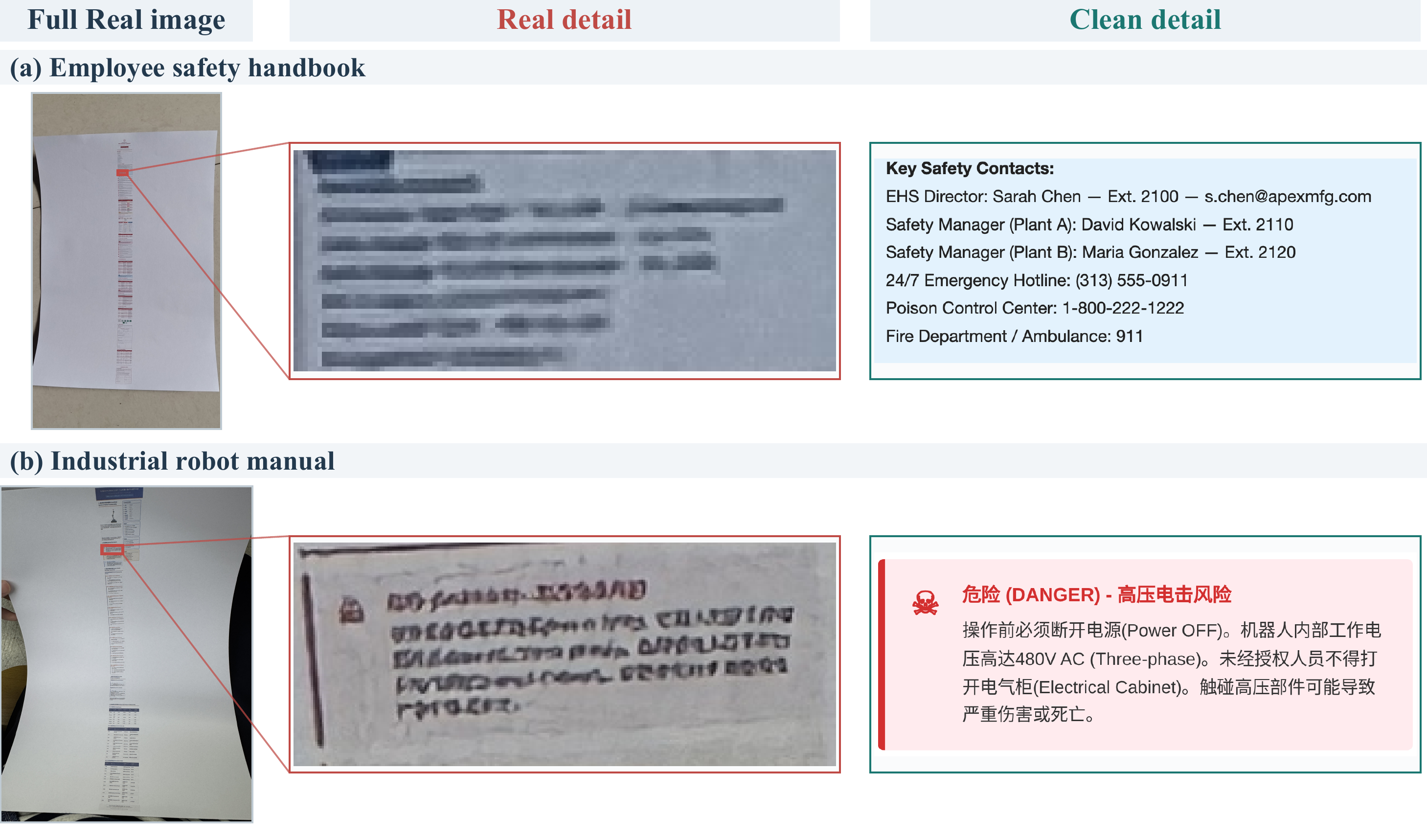}
\caption{\textbf{Readability limits in Real recaptures of long pages.} Fine text remains readable in Clean but loses essential detail during recapture, obscuring emergency contacts and a high-voltage warning. Both Real images belong to the 83 pages identified as unreadable in human review.}
\label{fig:readability_evidence}
\end{figure}

\paragraph{Screening and human review.} We analyzed the geometry of all 1,475 base pages. A hypothetical resize to a longest side of 1,024 pixels flagged 492 candidates whose effective width would fall below 512 pixels; 104 fell below 256 pixels. These thresholds prioritize long pages for inspection. Two annotators independently reviewed each candidate's Clean, Digital, and Real images at original resolution with zoom and reconciled their decisions. The 1,024-pixel setting is a geometric screening diagnostic; the final judgments use original-resolution content. The audit identified 83 Real images (5.63\%) whose text could not be reliably verified against its transcription. Their Clean and Digital counterparts remained readable. Figure~\ref{fig:readability_evidence} illustrates how fitting a long page into a recaptured image can remove fine textual detail. The released page-level decision list records image identifiers, geometry, track-specific readability, and the final decision.

\paragraph{Sensitivity protocol.} Table~\ref{tab:readability_sensitivity} compares full and filtered evaluation for 51 models. The filtered Real set excludes the 83 flagged images, retaining 1,392 pages. Clean and Digital retain all 1,475 pages and their original scores; each model's Avg$_3$ is recomputed with its filtered Real score. This analysis measures the effect of the Real exclusion on aggregate scores and model rankings. Both rankings are computed within the same 51-model cohort. The full-track results in Table~\ref{tab:leaderboard} retain their original evaluation scope.

\paragraph{Effect on scores and rankings.} Exclusion increases every model's Real score, with a mean gain of \ReadabilityRealMeanGain{} points and an Avg$_3$ gain of \ReadabilityAvgMeanGain{} points. Spearman correlations are $\rho{=}\ReadabilityRealSpearman$ for Real and $\rho{=}\ReadabilityAvgSpearman$ for Avg$_3$; the maximum absolute rank change is two positions in both cases. Both top-ten sets remain unchanged, as do the Real leader (Gemini~3.6 Flash) and Avg$_3$ leader (GLM-5.3-Flash). Thus, acquisition-related legibility loss depresses absolute scores, while relative rankings remain stable within this cohort. This motivates improved capture and readability screening alongside evaluation on difficult but legible inputs.

\clearpage
\begin{table}[H]
\centering
\caption{\textbf{Readability sensitivity for 51 models.} Removing 83 unreadable Real images preserves top-ten membership. Full and Filtered use 1,475 and 1,392 Real pages; Clean and Digital stay fixed. Avg$_3$ averages the three tracks, and $\Delta$ is Filtered minus Full. \textbf{Bold}/\underline{underline}: best/runner-up.}
\label{tab:readability_sensitivity}
{\fontsize{9}{10}\selectfont
\definecolor{sensitivityGain}{HTML}{167653}
\definecolor{sensitivityDrop}{HTML}{B52F3F}
\newlength{\sensitivityscorewidth}
\settowidth{\sensitivityscorewidth}{00.00}
\newlength{\sensitivitydeltawidth}
\settowidth{\sensitivitydeltawidth}{+0.00}
\newcommand{\sensitivityscoreheader}[1]{\makebox[\sensitivityscorewidth][c]{\textbf{#1}}}
\newcommand{\sensitivitydeltaheader}{\makebox[\sensitivitydeltawidth][c]{$\Delta$}}
\setlength{\tabcolsep}{2pt}
\renewcommand{\arraystretch}{0.96}
\begin{tabular}{>{\raggedright\arraybackslash}p{\dimexpr\textwidth-200pt-14\tabcolsep\relax}*{2}{>{\raggedleft\arraybackslash}p{32pt}>{\raggedleft\arraybackslash}p{36pt}>{\raggedleft\arraybackslash}p{32pt}}}
\toprule
\textbf{Model} & \multicolumn{3}{c}{\cellcolor{realColumn}\textcolor{realInk}{\textbf{Real}$\uparrow$}} & \multicolumn{3}{c}{\cellcolor{cleanColumn}\textcolor{cleanInk}{\textbf{Avg$_3$}$\uparrow$}} \\
\cmidrule(lr){2-4}\cmidrule(lr){5-7}
 & \sensitivityscoreheader{Full} & \sensitivityscoreheader{Filtered} & \sensitivitydeltaheader & \sensitivityscoreheader{Full} & \sensitivityscoreheader{Filtered} & \sensitivitydeltaheader \\
\midrule
\rowcolor{pipelineBand}\multicolumn{7}{l}{\strut\textbf{Pipeline / multi-stage specialists}} \\
\rowcolor{resultStripe}
\modelentry{paddleocr-vl-1-6-layout-corrected} & 54.24 & 56.08 & \textcolor{sensitivityGain}{+1.83} & 59.33 & 59.94 & \textcolor{sensitivityGain}{+0.61} \\
\modelentry{youtu-parsing} & 60.29 & 63.26 & \textcolor{sensitivityGain}{+2.97} & 68.32 & 69.31 & \textcolor{sensitivityGain}{+0.99} \\
\rowcolor{resultStripe}
\modelentry{dotsmocr} & 61.73 & 64.57 & \textcolor{sensitivityGain}{+2.84} & 70.39 & 71.33 & \textcolor{sensitivityGain}{+0.95} \\
\modelentry{glm-ocr} & 58.31 & 60.82 & \textcolor{sensitivityGain}{+2.52} & 63.34 & 64.18 & \textcolor{sensitivityGain}{+0.84} \\
\rowcolor{resultStripe}
\modelentry{mineru2-5-pro} & 62.56 & 65.36 & \textcolor{sensitivityGain}{+2.80} & 70.07 & 71.00 & \textcolor{sensitivityGain}{+0.93} \\
\modelentry{mineru2-5} & 59.15 & 61.94 & \textcolor{sensitivityGain}{+2.80} & 67.66 & 68.59 & \textcolor{sensitivityGain}{+0.93} \\
\rowcolor{resultStripe}
\modelentry{monkeyocr-v2-b-parsing} & 58.83 & 60.80 & \textcolor{sensitivityGain}{+1.97} & 64.91 & 65.57 & \textcolor{sensitivityGain}{+0.66} \\
\modelentry{monkeyocr-pro-3b} & 46.49 & 48.76 & \textcolor{sensitivityGain}{+2.27} & 55.37 & 56.13 & \textcolor{sensitivityGain}{+0.76} \\
\rowcolor{resultStripe}
\modelentry{monkeyocr-pro-1-2b} & 43.82 & 46.11 & \textcolor{sensitivityGain}{+2.29} & 53.54 & 54.31 & \textcolor{sensitivityGain}{+0.76} \\
\modelentry{openocr} & 25.73 & 27.02 & \textcolor{sensitivityGain}{+1.29} & 29.49 & 29.92 & \textcolor{sensitivityGain}{+0.43} \\
\rowcolor{resultStripe}
\modelentry{dolphin-v2} & 44.92 & 47.23 & \textcolor{sensitivityGain}{+2.31} & 57.02 & 57.79 & \textcolor{sensitivityGain}{+0.77} \\
\modelentry{teleocr} & 69.08 & 71.60 & \textcolor{sensitivityGain}{+2.52} & \underline{78.55} & \underline{79.39} & \textcolor{sensitivityGain}{+0.84} \\
\addlinespace[2pt]
\rowcolor{e2eBand}\multicolumn{7}{l}{\strut\textbf{End-to-end specialists}} \\
\rowcolor{resultStripe}
\modelentry{ovisocr2} & 66.49 & 69.22 & \textcolor{sensitivityGain}{+2.73} & 75.10 & 76.01 & \textcolor{sensitivityGain}{+0.91} \\
\modelentry{abot-ocr} & 61.77 & 64.05 & \textcolor{sensitivityGain}{+2.29} & 70.37 & 71.13 & \textcolor{sensitivityGain}{+0.76} \\
\rowcolor{resultStripe}
\modelentry{qianfan-ocr} & 45.06 & 47.48 & \textcolor{sensitivityGain}{+2.42} & 51.04 & 51.85 & \textcolor{sensitivityGain}{+0.81} \\
\modelentry{unlimited-ocr} & 53.71 & 56.06 & \textcolor{sensitivityGain}{+2.34} & 63.69 & 64.47 & \textcolor{sensitivityGain}{+0.78} \\
\rowcolor{resultStripe}
\modelentry{wevisdoc-4b} & 69.14 & 71.41 & \textcolor{sensitivityGain}{+2.27} & 75.57 & 76.33 & \textcolor{sensitivityGain}{+0.76} \\
\modelentry{wevisdoc-2b} & 65.68 & 68.11 & \textcolor{sensitivityGain}{+2.43} & 73.69 & 74.50 & \textcolor{sensitivityGain}{+0.81} \\
\rowcolor{resultStripe}
\modelentry{hunyuanocr-1-5} & 60.28 & 62.72 & \textcolor{sensitivityGain}{+2.44} & 67.58 & 68.39 & \textcolor{sensitivityGain}{+0.81} \\
\modelentry{dots-ocr} & 55.68 & 58.69 & \textcolor{sensitivityGain}{+3.01} & 64.55 & 65.55 & \textcolor{sensitivityGain}{+1.00} \\
\rowcolor{resultStripe}
\modelentry{firered-ocr} & 57.42 & 60.16 & \textcolor{sensitivityGain}{+2.75} & 65.57 & 66.49 & \textcolor{sensitivityGain}{+0.92} \\
\modelentry{deepseek-ocr} & 40.48 & 42.76 & \textcolor{sensitivityGain}{+2.28} & 46.98 & 47.74 & \textcolor{sensitivityGain}{+0.76} \\
\rowcolor{resultStripe}
\modelentry{olmocr-2-7b} & 56.10 & 58.90 & \textcolor{sensitivityGain}{+2.80} & 63.78 & 64.71 & \textcolor{sensitivityGain}{+0.93} \\
\modelentry{olmocr-7b} & 47.30 & 49.64 & \textcolor{sensitivityGain}{+2.34} & 55.90 & 56.68 & \textcolor{sensitivityGain}{+0.78} \\
\rowcolor{resultStripe}
\modelentry{ocrverse} & 63.66 & 66.25 & \textcolor{sensitivityGain}{+2.59} & 69.40 & 70.26 & \textcolor{sensitivityGain}{+0.86} \\
\modelentry{fd-rl} & 67.04 & 69.71 & \textcolor{sensitivityGain}{+2.67} & 73.92 & 74.81 & \textcolor{sensitivityGain}{+0.89} \\
\rowcolor{resultStripe}
\modelentry{nanonets-ocr2} & 49.03 & 51.70 & \textcolor{sensitivityGain}{+2.67} & 58.36 & 59.25 & \textcolor{sensitivityGain}{+0.89} \\
\modelentry{ocrflux-3b} & 37.21 & 39.05 & \textcolor{sensitivityGain}{+1.83} & 42.06 & 42.67 & \textcolor{sensitivityGain}{+0.61} \\
\rowcolor{resultStripe}
\modelentry{logics-parsing-v2} & 67.64 & 70.21 & \textcolor{sensitivityGain}{+2.56} & 72.61 & 73.47 & \textcolor{sensitivityGain}{+0.85} \\
\addlinespace[2pt]
\rowcolor{vlmBand}\multicolumn{7}{l}{\strut\textbf{General-purpose VLMs}} \\
\rowcolor{resultStripe}
\modelentry{qwen3-8-flash-next} & 62.42 & 64.57 & \textcolor{sensitivityGain}{+2.15} & 68.54 & 69.25 & \textcolor{sensitivityGain}{+0.72} \\
\modelentry{qwen3-6-35b-a3b} & 60.12 & 62.16 & \textcolor{sensitivityGain}{+2.04} & 67.14 & 67.82 & \textcolor{sensitivityGain}{+0.68} \\
\rowcolor{resultStripe}
\modelentry{qwen3-6-27b} & 59.18 & 61.18 & \textcolor{sensitivityGain}{+2.00} & 65.62 & 66.29 & \textcolor{sensitivityGain}{+0.67} \\
\modelentry{qwen3-5-397b-a17b} & 62.70 & 64.60 & \textcolor{sensitivityGain}{+1.90} & 66.72 & 67.35 & \textcolor{sensitivityGain}{+0.63} \\
\rowcolor{resultStripe}
\modelentry{qwen3-5-122b-a10b} & 69.85 & 72.21 & \textcolor{sensitivityGain}{+2.36} & 74.11 & 74.90 & \textcolor{sensitivityGain}{+0.79} \\
\modelentry{qwen3-5-35b-a3b} & 60.59 & 62.59 & \textcolor{sensitivityGain}{+2.01} & 65.68 & 66.34 & \textcolor{sensitivityGain}{+0.67} \\
\rowcolor{resultStripe}
\modelentry{qwen3-5-27b} & 65.92 & 68.13 & \textcolor{sensitivityGain}{+2.21} & 69.57 & 70.31 & \textcolor{sensitivityGain}{+0.74} \\
\modelentry{qwen3-5-9b} & 65.45 & 67.76 & \textcolor{sensitivityGain}{+2.32} & 70.89 & 71.66 & \textcolor{sensitivityGain}{+0.77} \\
\rowcolor{resultStripe}
\modelentry{qwen3-5-4b} & 63.47 & 66.02 & \textcolor{sensitivityGain}{+2.56} & 69.82 & 70.67 & \textcolor{sensitivityGain}{+0.85} \\
\modelentry{qwen3-5-2b} & 55.92 & 58.39 & \textcolor{sensitivityGain}{+2.47} & 62.46 & 63.28 & \textcolor{sensitivityGain}{+0.82} \\
\rowcolor{resultStripe}
\modelentry{qwen3-5-0-8b} & 47.93 & 50.56 & \textcolor{sensitivityGain}{+2.62} & 55.99 & 56.87 & \textcolor{sensitivityGain}{+0.87} \\
\modelentry{qwen3-vl-8b} & 62.73 & 65.31 & \textcolor{sensitivityGain}{+2.58} & 69.07 & 69.93 & \textcolor{sensitivityGain}{+0.86} \\
\rowcolor{resultStripe}
\modelentry{qwen3-vl-4b} & 59.61 & 62.30 & \textcolor{sensitivityGain}{+2.69} & 67.50 & 68.39 & \textcolor{sensitivityGain}{+0.90} \\
\modelentry{qwen3-vl-2b} & 54.09 & 56.60 & \textcolor{sensitivityGain}{+2.50} & 62.09 & 62.93 & \textcolor{sensitivityGain}{+0.83} \\
\rowcolor{resultStripe}
\modelentry{deepseek-v4-flash-vision-exp} & 42.68 & 44.09 & \textcolor{sensitivityGain}{+1.42} & 50.12 & 50.59 & \textcolor{sensitivityGain}{+0.47} \\
\modelentry{glm-5-3-flash} & \underline{74.99} & \underline{77.02} & \textcolor{sensitivityGain}{+2.03} & \textbf{79.69} & \textbf{80.36} & \textcolor{sensitivityGain}{+0.68} \\
\rowcolor{resultStripe}
\modelentry{minicpm-v-4-6} & 48.01 & 49.85 & \textcolor{sensitivityGain}{+1.84} & 57.17 & 57.79 & \textcolor{sensitivityGain}{+0.61} \\
\modelentry{minicpm-v-4-5} & 37.59 & 39.54 & \textcolor{sensitivityGain}{+1.95} & 46.26 & 46.91 & \textcolor{sensitivityGain}{+0.65} \\
\rowcolor{resultStripe}
\modelentry{gemini-3-6-flash} & \textbf{75.73} & \textbf{77.30} & \textcolor{sensitivityGain}{+1.57} & 78.27 & 78.79 & \textcolor{sensitivityGain}{+0.52} \\
\modelentry{gemini-3-1-pro} & 71.98 & 74.03 & \textcolor{sensitivityGain}{+2.05} & 70.43 & 71.12 & \textcolor{sensitivityGain}{+0.68} \\
\rowcolor{resultStripe}
\modelentry{kimi-k2-6} & 68.02 & 70.56 & \textcolor{sensitivityGain}{+2.54} & 70.10 & 70.94 & \textcolor{sensitivityGain}{+0.85} \\
\modelentry{step3-vl} & 45.06 & 46.96 & \textcolor{sensitivityGain}{+1.90} & 50.48 & 51.12 & \textcolor{sensitivityGain}{+0.63} \\
\bottomrule
\end{tabular}
}
\end{table}

\clearpage
\section{Full Component Results}
\label{app:full_results}
{\fontsize{9}{9.5}\selectfont
\setlength{\tabcolsep}{4pt}
\renewcommand{\arraystretch}{0.94}
\setlength{\LTleft}{0pt}
\setlength{\LTright}{0pt}
\setlength{\LTcapwidth}{\textwidth}
\begin{longtable}{>{\raggedright\arraybackslash}p{\dimexpr\textwidth-175pt-12\tabcolsep\relax}*{5}{>{\raggedleft\arraybackslash}p{35pt}}}
\caption{\textbf{Clean component results for 58 models.} Text and Reading Edit are normalized edit distances ($\downarrow$); Formula CDM, Table TEDS, and Overall use a shared 0--100 scoring scale ($\uparrow$). \textbf{Bold}/\underline{underline}: best/runner-up.}\label{tab:components_clean}\\
\toprule
\textbf{Model} & \textbf{Overall}$\uparrow$ & \makecell{\textbf{Text}\\\textbf{Edit}$\downarrow$} & \makecell{\textbf{Formula}\\\textbf{CDM}$\uparrow$} & \makecell{\textbf{Table}\\\textbf{TEDS}$\uparrow$} & \makecell{\textbf{Reading}\\\textbf{Edit}$\downarrow$} \\
\midrule
\endfirsthead
\multicolumn{6}{l}{\textit{Clean component results (continued)}}\\
\toprule
\textbf{Model} & \textbf{Overall}$\uparrow$ & \makecell{\textbf{Text}\\\textbf{Edit}$\downarrow$} & \makecell{\textbf{Formula}\\\textbf{CDM}$\uparrow$} & \makecell{\textbf{Table}\\\textbf{TEDS}$\uparrow$} & \makecell{\textbf{Reading}\\\textbf{Edit}$\downarrow$} \\
\midrule
\endhead
\bottomrule
\endfoot
\rowcolor{pipelineBand}\multicolumn{6}{l}{\strut\textbf{Pipeline / multi-stage specialists}} \\
\rowcolor{resultStripe}
\modelentry{paddleocr-vl-1-6-layout-corrected} & 64.46 & 0.432 & 53.91 & 82.70 & 0.556 \\
\modelentry{youtu-parsing} & 75.02 & 0.230 & 67.34 & 80.74 & 0.358 \\
\rowcolor{resultStripe}
\modelentry{dotsmocr} & 76.27 & \underline{0.151} & 66.23 & 77.65 & \underline{0.273} \\
\modelentry{glm-ocr} & 68.65 & 0.314 & 57.89 & 79.44 & 0.470 \\
\rowcolor{resultStripe}
\modelentry{mineru2-5-pro} & 75.87 & 0.222 & 65.14 & 84.68 & 0.346 \\
\modelentry{mineru2-5} & 74.90 & 0.184 & 62.08 & 81.04 & 0.327 \\
\rowcolor{resultStripe}
\modelentry{monkeyocr-v2-b-parsing} & 69.82 & 0.328 & 62.27 & 80.05 & 0.409 \\
\modelentry{monkeyocr-pro-3b} & 62.23 & 0.346 & 48.46 & 72.83 & 0.492 \\
\rowcolor{resultStripe}
\modelentry{monkeyocr-pro-1-2b} & 61.09 & 0.358 & 47.43 & 71.60 & 0.498 \\
\modelentry{opendoc-0-1b} & 60.28 & 0.411 & 53.09 & 68.86 & 0.519 \\
\rowcolor{resultStripe}
\modelentry{openocr} & 32.70 & 0.354 & 33.50 & 0.00 & 0.507 \\
\modelentry{dolphin-v2} & 65.90 & 0.342 & 59.80 & 72.12 & 0.429 \\
\rowcolor{resultStripe}
\modelentry{teleocr} & \textbf{87.16} & \textbf{0.120} & \textbf{82.33} & \underline{91.12} & \textbf{0.157} \\
\addlinespace[2pt]
\rowcolor{e2eBand}\multicolumn{6}{l}{\strut\textbf{End-to-end specialists}} \\
\rowcolor{resultStripe}
\modelentry{ovisocr2} & 81.53 & 0.152 & 69.66 & 90.10 & 0.282 \\
\modelentry{abot-ocr} & 76.92 & 0.224 & 66.81 & 86.39 & 0.397 \\
\rowcolor{resultStripe}
\modelentry{qianfan-ocr} & 57.22 & 0.370 & 49.79 & 58.83 & 0.443 \\
\modelentry{unlimited-ocr} & 72.05 & 0.243 & 63.68 & 76.81 & 0.367 \\
\rowcolor{resultStripe}
\modelentry{wevisdoc-4b} & 79.59 & 0.204 & 69.95 & 89.23 & 0.332 \\
\modelentry{wevisdoc-2b} & 79.08 & 0.217 & 71.25 & 87.64 & 0.347 \\
\rowcolor{resultStripe}
\modelentry{hunyuanocr-1-5} & 73.09 & 0.242 & 67.54 & 75.94 & 0.391 \\
\modelentry{hunyuanocr} & 65.61 & 0.269 & 55.74 & 68.02 & 0.382 \\
\rowcolor{resultStripe}
\modelentry{dots-ocr} & 72.01 & 0.248 & 61.37 & 79.51 & 0.379 \\
\modelentry{firered-ocr} & 70.81 & 0.287 & 63.86 & 77.23 & 0.396 \\
\rowcolor{resultStripe}
\modelentry{deepseek-ocr-2} & 55.53 & 0.354 & 46.00 & 56.01 & 0.466 \\
\modelentry{deepseek-ocr} & 53.50 & 0.419 & 45.39 & 57.06 & 0.514 \\
\rowcolor{resultStripe}
\modelentry{olmocr-2-7b} & 69.36 & 0.284 & 56.89 & 79.59 & 0.358 \\
\modelentry{olmocr-7b} & 62.56 & 0.388 & 58.69 & 67.77 & 0.466 \\
\rowcolor{resultStripe}
\modelentry{ocrverse} & 73.18 & 0.273 & 63.78 & 83.09 & 0.393 \\
\modelentry{fd-rl} & 78.38 & 0.193 & 68.21 & 86.22 & 0.334 \\
\rowcolor{resultStripe}
\modelentry{nanonets-ocr2} & 64.83 & 0.254 & 44.98 & 74.94 & 0.377 \\
\modelentry{ocrflux-3b} & 47.14 & 0.454 & 38.35 & 48.46 & 0.424 \\
\rowcolor{resultStripe}
\modelentry{logics-parsing-v2} & 76.35 & 0.213 & 67.67 & 82.67 & 0.342 \\
\addlinespace[2pt]
\rowcolor{vlmBand}\multicolumn{6}{l}{\strut\textbf{General-purpose VLMs}} \\
\rowcolor{resultStripe}
\modelentry{qwen3-8-27b} & 80.45 & 0.180 & 71.68 & 87.69 & 0.305 \\
\modelentry{qwen3-8-flash-next} & 72.23 & 0.325 & 72.28 & 76.93 & 0.420 \\
\rowcolor{resultStripe}
\modelentry{qwen3-6-35b-a3b} & 72.14 & 0.302 & 70.39 & 76.25 & 0.397 \\
\modelentry{qwen3-6-27b} & 70.20 & 0.352 & 69.36 & 76.47 & 0.446 \\
\rowcolor{resultStripe}
\modelentry{qwen3-5-397b-a17b} & 69.12 & 0.233 & 65.26 & 65.40 & 0.366 \\
\modelentry{qwen3-5-122b-a10b} & 76.14 & 0.226 & 67.96 & 83.03 & 0.375 \\
\rowcolor{resultStripe}
\modelentry{qwen3-5-35b-a3b} & 68.40 & 0.232 & 64.94 & 63.45 & 0.374 \\
\modelentry{qwen3-5-27b} & 72.07 & 0.227 & 66.36 & 72.51 & 0.362 \\
\rowcolor{resultStripe}
\modelentry{qwen3-5-9b} & 73.87 & 0.254 & 67.60 & 79.39 & 0.388 \\
\modelentry{qwen3-5-4b} & 73.45 & 0.276 & 69.96 & 78.02 & 0.410 \\
\rowcolor{resultStripe}
\modelentry{qwen3-5-2b} & 66.24 & 0.348 & 62.84 & 70.70 & 0.473 \\
\modelentry{qwen3-5-0-8b} & 60.77 & 0.376 & 54.39 & 65.54 & 0.500 \\
\rowcolor{resultStripe}
\modelentry{qwen3-vl-8b} & 72.44 & 0.261 & 65.10 & 78.35 & 0.411 \\
\modelentry{qwen3-vl-4b} & 72.04 & 0.262 & 65.10 & 77.17 & 0.418 \\
\rowcolor{resultStripe}
\modelentry{qwen3-vl-2b} & 66.37 & 0.300 & 59.04 & 70.03 & 0.439 \\
\modelentry{deepseek-v4-flash-vision-exp} & 54.69 & 0.540 & 57.47 & 60.55 & 0.567 \\
\rowcolor{resultStripe}
\modelentry{glm-5-3-flash} & \underline{83.04} & 0.218 & \underline{80.16} & 90.76 & 0.375 \\
\modelentry{minicpm-v-4-6} & 64.47 & 0.343 & 58.05 & 69.72 & 0.437 \\
\rowcolor{resultStripe}
\modelentry{minicpm-v-4-5} & 51.81 & 0.439 & 45.97 & 53.36 & 0.481 \\
\modelentry{gemini-3-6-flash} & 79.97 & 0.251 & 78.29 & 86.76 & 0.381 \\
\rowcolor{resultStripe}
\modelentry{gemini-3-1-pro} & 70.04 & 0.306 & 65.63 & 75.08 & 0.409 \\
\modelentry{claude-opus-5} & 82.99 & 0.191 & 75.92 & \textbf{92.17} & 0.336 \\
\rowcolor{resultStripe}
\modelentry{gpt-5-6-sol} & 74.15 & 0.354 & 74.96 & 82.87 & 0.440 \\
\modelentry{minimax-m3} & 55.78 & 0.506 & 54.41 & 63.53 & 0.529 \\
\rowcolor{resultStripe}
\modelentry{kimi-k2-6} & 72.32 & 0.303 & 66.93 & 80.30 & 0.466 \\
\modelentry{step3-vl} & 53.65 & 0.496 & 53.41 & 57.16 & 0.509 \\
\end{longtable}
}
\clearpage
{\fontsize{9}{9.5}\selectfont
\setlength{\tabcolsep}{4pt}
\renewcommand{\arraystretch}{0.94}
\setlength{\LTleft}{0pt}
\setlength{\LTright}{0pt}
\setlength{\LTcapwidth}{\textwidth}
\begin{longtable}{>{\raggedright\arraybackslash}p{\dimexpr\textwidth-175pt-12\tabcolsep\relax}*{5}{>{\raggedleft\arraybackslash}p{35pt}}}
\caption{\textbf{Digital component results for 58 models.} Text and Reading Edit are normalized edit distances ($\downarrow$); Formula CDM, Table TEDS, and Overall use a shared 0--100 scoring scale ($\uparrow$). \textbf{Bold}/\underline{underline}: best/runner-up.}\label{tab:components_digital}\\
\toprule
\textbf{Model} & \textbf{Overall}$\uparrow$ & \makecell{\textbf{Text}\\\textbf{Edit}$\downarrow$} & \makecell{\textbf{Formula}\\\textbf{CDM}$\uparrow$} & \makecell{\textbf{Table}\\\textbf{TEDS}$\uparrow$} & \makecell{\textbf{Reading}\\\textbf{Edit}$\downarrow$} \\
\midrule
\endfirsthead
\multicolumn{6}{l}{\textit{Digital component results (continued)}}\\
\toprule
\textbf{Model} & \textbf{Overall}$\uparrow$ & \makecell{\textbf{Text}\\\textbf{Edit}$\downarrow$} & \makecell{\textbf{Formula}\\\textbf{CDM}$\uparrow$} & \makecell{\textbf{Table}\\\textbf{TEDS}$\uparrow$} & \makecell{\textbf{Reading}\\\textbf{Edit}$\downarrow$} \\
\midrule
\endhead
\bottomrule
\endfoot
\rowcolor{pipelineBand}\multicolumn{6}{l}{\strut\textbf{Pipeline / multi-stage specialists}} \\
\rowcolor{resultStripe}
\modelentry{paddleocr-vl-1-6-layout-corrected} & 59.29 & 0.494 & 51.78 & 75.46 & 0.596 \\
\modelentry{youtu-parsing} & 69.66 & 0.270 & 61.44 & 74.49 & 0.388 \\
\rowcolor{resultStripe}
\modelentry{dotsmocr} & 73.16 & \underline{0.198} & 64.32 & 74.95 & \underline{0.309} \\
\modelentry{glm-ocr} & 63.06 & 0.383 & 53.23 & 74.21 & 0.520 \\
\rowcolor{resultStripe}
\modelentry{mineru2-5-pro} & 71.77 & 0.272 & 61.79 & 80.73 & 0.378 \\
\modelentry{mineru2-5} & 68.92 & 0.245 & 56.99 & 74.24 & 0.374 \\
\rowcolor{resultStripe}
\modelentry{monkeyocr-v2-b-parsing} & 66.08 & 0.356 & 59.06 & 74.74 & 0.426 \\
\modelentry{monkeyocr-pro-3b} & 57.40 & 0.397 & 45.57 & 66.32 & 0.526 \\
\rowcolor{resultStripe}
\modelentry{monkeyocr-pro-1-2b} & 55.72 & 0.416 & 43.91 & 64.83 & 0.529 \\
\modelentry{opendoc-0-1b} & 52.46 & 0.501 & 48.41 & 59.04 & 0.577 \\
\rowcolor{resultStripe}
\modelentry{openocr} & 30.03 & 0.410 & 31.09 & 0.00 & 0.541 \\
\modelentry{dolphin-v2} & 60.24 & 0.393 & 52.20 & 67.86 & 0.461 \\
\rowcolor{resultStripe}
\modelentry{teleocr} & \underline{79.40} & 0.209 & 76.12 & 82.94 & \textbf{0.237} \\
\addlinespace[2pt]
\rowcolor{e2eBand}\multicolumn{6}{l}{\strut\textbf{End-to-end specialists}} \\
\rowcolor{resultStripe}
\modelentry{ovisocr2} & 77.29 & \textbf{0.196} & 66.49 & 84.94 & 0.316 \\
\modelentry{abot-ocr} & 72.43 & 0.269 & 63.16 & 80.99 & 0.435 \\
\rowcolor{resultStripe}
\modelentry{qianfan-ocr} & 50.85 & 0.438 & 44.41 & 51.96 & 0.485 \\
\modelentry{unlimited-ocr} & 65.30 & 0.310 & 57.63 & 69.32 & 0.412 \\
\rowcolor{resultStripe}
\modelentry{wevisdoc-4b} & 77.99 & 0.234 & 70.75 & \underline{86.65} & 0.356 \\
\modelentry{wevisdoc-2b} & 76.33 & 0.255 & 70.34 & 84.09 & 0.374 \\
\rowcolor{resultStripe}
\modelentry{hunyuanocr-1-5} & 69.36 & 0.275 & 64.63 & 70.99 & 0.414 \\
\modelentry{hunyuanocr} & 61.49 & 0.308 & 51.62 & 63.68 & 0.400 \\
\rowcolor{resultStripe}
\modelentry{dots-ocr} & 65.95 & 0.307 & 56.67 & 71.86 & 0.417 \\
\modelentry{firered-ocr} & 68.49 & 0.319 & 62.64 & 74.77 & 0.422 \\
\rowcolor{resultStripe}
\modelentry{deepseek-ocr-2} & 49.41 & 0.412 & 40.78 & 48.67 & 0.493 \\
\modelentry{deepseek-ocr} & 46.95 & 0.478 & 39.99 & 48.64 & 0.548 \\
\rowcolor{resultStripe}
\modelentry{olmocr-2-7b} & 65.87 & 0.318 & 54.57 & 74.81 & 0.378 \\
\modelentry{olmocr-7b} & 57.84 & 0.436 & 55.44 & 61.66 & 0.499 \\
\rowcolor{resultStripe}
\modelentry{ocrverse} & 71.36 & 0.302 & 63.95 & 80.36 & 0.415 \\
\modelentry{fd-rl} & 76.33 & 0.214 & 67.16 & 83.22 & 0.350 \\
\rowcolor{resultStripe}
\modelentry{nanonets-ocr2} & 61.23 & 0.307 & 45.40 & 68.97 & 0.408 \\
\modelentry{ocrflux-3b} & 41.82 & 0.486 & 31.90 & 42.17 & 0.437 \\
\rowcolor{resultStripe}
\modelentry{logics-parsing-v2} & 73.85 & 0.248 & 67.33 & 79.02 & 0.375 \\
\addlinespace[2pt]
\rowcolor{vlmBand}\multicolumn{6}{l}{\strut\textbf{General-purpose VLMs}} \\
\rowcolor{resultStripe}
\modelentry{qwen3-8-27b} & 78.86 & 0.206 & 71.98 & 85.15 & 0.325 \\
\modelentry{qwen3-8-flash-next} & 70.96 & 0.346 & 72.87 & 74.59 & 0.430 \\
\rowcolor{resultStripe}
\modelentry{qwen3-6-35b-a3b} & 69.16 & 0.329 & 67.67 & 72.66 & 0.412 \\
\modelentry{qwen3-6-27b} & 67.49 & 0.375 & 66.54 & 73.42 & 0.457 \\
\rowcolor{resultStripe}
\modelentry{qwen3-5-397b-a17b} & 68.34 & 0.244 & 63.91 & 65.53 & 0.376 \\
\modelentry{qwen3-5-122b-a10b} & 76.34 & 0.220 & 67.82 & 83.21 & 0.366 \\
\rowcolor{resultStripe}
\modelentry{qwen3-5-35b-a3b} & 68.04 & 0.245 & 64.78 & 63.86 & 0.379 \\
\modelentry{qwen3-5-27b} & 70.73 & 0.236 & 64.61 & 71.17 & 0.367 \\
\rowcolor{resultStripe}
\modelentry{qwen3-5-9b} & 73.34 & 0.260 & 67.00 & 79.01 & 0.396 \\
\modelentry{qwen3-5-4b} & 72.53 & 0.281 & 68.88 & 76.78 & 0.412 \\
\rowcolor{resultStripe}
\modelentry{qwen3-5-2b} & 65.22 & 0.350 & 58.30 & 72.36 & 0.477 \\
\modelentry{qwen3-5-0-8b} & 59.28 & 0.386 & 54.22 & 62.22 & 0.510 \\
\rowcolor{resultStripe}
\modelentry{qwen3-vl-8b} & 72.03 & 0.266 & 64.88 & 77.82 & 0.409 \\
\modelentry{qwen3-vl-4b} & 70.84 & 0.272 & 63.54 & 76.13 & 0.425 \\
\rowcolor{resultStripe}
\modelentry{qwen3-vl-2b} & 65.81 & 0.314 & 60.25 & 68.52 & 0.448 \\
\modelentry{deepseek-v4-flash-vision-exp} & 52.98 & 0.561 & 57.24 & 57.77 & 0.576 \\
\rowcolor{resultStripe}
\modelentry{glm-5-3-flash} & \textbf{81.03} & 0.242 & \underline{78.51} & \textbf{88.84} & 0.391 \\
\modelentry{minicpm-v-4-6} & 59.04 & 0.395 & 54.59 & 62.00 & 0.473 \\
\rowcolor{resultStripe}
\modelentry{minicpm-v-4-5} & 49.38 & 0.461 & 42.79 & 51.50 & 0.489 \\
\modelentry{gemini-3-6-flash} & 79.11 & 0.268 & \textbf{78.88} & 85.31 & 0.391 \\
\rowcolor{resultStripe}
\modelentry{gemini-3-1-pro} & 69.28 & 0.322 & 65.81 & 74.24 & 0.417 \\
\modelentry{claude-opus-5} & 72.94 & 0.351 & 73.43 & 80.45 & 0.420 \\
\rowcolor{resultStripe}
\modelentry{gpt-5-6-sol} & 71.73 & 0.380 & 72.83 & 80.33 & 0.447 \\
\modelentry{minimax-m3} & 53.85 & 0.532 & 54.15 & 60.55 & 0.548 \\
\rowcolor{resultStripe}
\modelentry{kimi-k2-6} & 69.95 & 0.322 & 64.69 & 77.31 & 0.475 \\
\modelentry{step3-vl} & 52.74 & 0.516 & 53.62 & 56.15 & 0.529 \\
\end{longtable}
}
\clearpage
{\fontsize{9}{9.5}\selectfont
\setlength{\tabcolsep}{4pt}
\renewcommand{\arraystretch}{0.94}
\setlength{\LTleft}{0pt}
\setlength{\LTright}{0pt}
\setlength{\LTcapwidth}{\textwidth}
\begin{longtable}{>{\raggedright\arraybackslash}p{\dimexpr\textwidth-175pt-12\tabcolsep\relax}*{5}{>{\raggedleft\arraybackslash}p{35pt}}}
\caption{\textbf{Real component results for 58 models.} Text and Reading Edit are normalized edit distances ($\downarrow$); Formula CDM, Table TEDS, and Overall use a shared 0--100 scoring scale ($\uparrow$). \textbf{Bold}/\underline{underline}: best/runner-up.}\label{tab:components_real}\\
\toprule
\textbf{Model} & \textbf{Overall}$\uparrow$ & \makecell{\textbf{Text}\\\textbf{Edit}$\downarrow$} & \makecell{\textbf{Formula}\\\textbf{CDM}$\uparrow$} & \makecell{\textbf{Table}\\\textbf{TEDS}$\uparrow$} & \makecell{\textbf{Reading}\\\textbf{Edit}$\downarrow$} \\
\midrule
\endfirsthead
\multicolumn{6}{l}{\textit{Real component results (continued)}}\\
\toprule
\textbf{Model} & \textbf{Overall}$\uparrow$ & \makecell{\textbf{Text}\\\textbf{Edit}$\downarrow$} & \makecell{\textbf{Formula}\\\textbf{CDM}$\uparrow$} & \makecell{\textbf{Table}\\\textbf{TEDS}$\uparrow$} & \makecell{\textbf{Reading}\\\textbf{Edit}$\downarrow$} \\
\midrule
\endhead
\bottomrule
\endfoot
\rowcolor{pipelineBand}\multicolumn{6}{l}{\strut\textbf{Pipeline / multi-stage specialists}} \\
\rowcolor{resultStripe}
\modelentry{paddleocr-vl-1-6-layout-corrected} & 54.24 & 0.529 & 43.90 & 71.76 & 0.619 \\
\modelentry{youtu-parsing} & 60.29 & 0.360 & 52.20 & 64.69 & 0.430 \\
\rowcolor{resultStripe}
\modelentry{dotsmocr} & 61.73 & 0.312 & 54.39 & 61.97 & 0.393 \\
\modelentry{glm-ocr} & 58.31 & 0.433 & 50.34 & 67.83 & 0.543 \\
\rowcolor{resultStripe}
\modelentry{mineru2-5-pro} & 62.56 & 0.375 & 52.70 & 72.47 & 0.446 \\
\modelentry{mineru2-5} & 59.15 & 0.370 & 49.01 & 65.41 & 0.446 \\
\rowcolor{resultStripe}
\modelentry{monkeyocr-v2-b-parsing} & 58.83 & 0.442 & 53.48 & 67.22 & 0.458 \\
\modelentry{monkeyocr-pro-3b} & 46.49 & 0.511 & 38.18 & 52.43 & 0.600 \\
\rowcolor{resultStripe}
\modelentry{monkeyocr-pro-1-2b} & 43.82 & 0.556 & 36.94 & 50.07 & 0.609 \\
\modelentry{opendoc-0-1b} & 44.27 & 0.547 & 38.46 & 49.06 & 0.603 \\
\rowcolor{resultStripe}
\modelentry{openocr} & 25.73 & 0.486 & 25.81 & 0.00 & 0.591 \\
\modelentry{dolphin-v2} & 44.92 & 0.553 & 39.98 & 50.04 & 0.558 \\
\rowcolor{resultStripe}
\modelentry{teleocr} & 69.08 & 0.338 & 64.69 & 76.32 & \textbf{0.352} \\
\addlinespace[2pt]
\rowcolor{e2eBand}\multicolumn{6}{l}{\strut\textbf{End-to-end specialists}} \\
\rowcolor{resultStripe}
\modelentry{ovisocr2} & 66.49 & 0.320 & 57.72 & 73.70 & 0.407 \\
\modelentry{abot-ocr} & 61.77 & 0.369 & 53.51 & 68.73 & 0.498 \\
\rowcolor{resultStripe}
\modelentry{qianfan-ocr} & 45.06 & 0.494 & 39.08 & 45.53 & 0.509 \\
\modelentry{unlimited-ocr} & 53.71 & 0.440 & 46.97 & 58.13 & 0.469 \\
\rowcolor{resultStripe}
\modelentry{wevisdoc-4b} & 69.14 & 0.333 & 62.74 & 77.92 & 0.411 \\
\modelentry{wevisdoc-2b} & 65.68 & 0.364 & 59.21 & 74.21 & 0.432 \\
\rowcolor{resultStripe}
\modelentry{hunyuanocr-1-5} & 60.28 & 0.376 & 56.62 & 61.81 & 0.476 \\
\modelentry{hunyuanocr} & 54.58 & 0.421 & 48.30 & 57.54 & 0.459 \\
\rowcolor{resultStripe}
\modelentry{dots-ocr} & 55.68 & 0.403 & 47.70 & 59.63 & 0.467 \\
\modelentry{firered-ocr} & 57.42 & 0.415 & 51.60 & 62.16 & 0.474 \\
\rowcolor{resultStripe}
\modelentry{deepseek-ocr-2} & 43.60 & 0.486 & 37.30 & 42.06 & 0.533 \\
\modelentry{deepseek-ocr} & 40.48 & 0.537 & 34.04 & 41.12 & 0.575 \\
\rowcolor{resultStripe}
\modelentry{olmocr-2-7b} & 56.10 & 0.417 & 48.79 & 61.25 & 0.439 \\
\modelentry{olmocr-7b} & 47.30 & 0.542 & 46.26 & 49.80 & 0.568 \\
\rowcolor{resultStripe}
\modelentry{ocrverse} & 63.66 & 0.363 & 57.03 & 70.30 & 0.452 \\
\modelentry{fd-rl} & 67.04 & 0.298 & 58.82 & 72.08 & 0.391 \\
\rowcolor{resultStripe}
\modelentry{nanonets-ocr2} & 49.03 & 0.435 & 35.50 & 55.09 & 0.468 \\
\modelentry{ocrflux-3b} & 37.21 & 0.559 & 32.65 & 34.87 & 0.491 \\
\rowcolor{resultStripe}
\modelentry{logics-parsing-v2} & 67.64 & 0.304 & 61.65 & 71.64 & 0.416 \\
\addlinespace[2pt]
\rowcolor{vlmBand}\multicolumn{6}{l}{\strut\textbf{General-purpose VLMs}} \\
\rowcolor{resultStripe}
\modelentry{qwen3-8-27b} & 73.59 & \textbf{0.274} & 68.56 & 79.63 & \underline{0.364} \\
\modelentry{qwen3-8-flash-next} & 62.42 & 0.424 & 62.84 & 66.85 & 0.472 \\
\rowcolor{resultStripe}
\modelentry{qwen3-6-35b-a3b} & 60.12 & 0.426 & 59.66 & 63.27 & 0.469 \\
\modelentry{qwen3-6-27b} & 59.18 & 0.453 & 56.51 & 66.34 & 0.504 \\
\rowcolor{resultStripe}
\modelentry{qwen3-5-397b-a17b} & 62.70 & 0.287 & 60.70 & 56.12 & 0.399 \\
\modelentry{qwen3-5-122b-a10b} & 69.85 & \underline{0.281} & 62.19 & 75.44 & 0.401 \\
\rowcolor{resultStripe}
\modelentry{qwen3-5-35b-a3b} & 60.59 & 0.310 & 59.68 & 53.07 & 0.419 \\
\modelentry{qwen3-5-27b} & 65.92 & 0.283 & 61.23 & 64.82 & 0.390 \\
\rowcolor{resultStripe}
\modelentry{qwen3-5-9b} & 65.45 & 0.332 & 60.91 & 68.59 & 0.437 \\
\modelentry{qwen3-5-4b} & 63.47 & 0.380 & 61.27 & 67.17 & 0.477 \\
\rowcolor{resultStripe}
\modelentry{qwen3-5-2b} & 55.92 & 0.440 & 50.99 & 60.79 & 0.521 \\
\modelentry{qwen3-5-0-8b} & 47.93 & 0.498 & 44.60 & 48.98 & 0.557 \\
\rowcolor{resultStripe}
\modelentry{qwen3-vl-8b} & 62.73 & 0.342 & 55.55 & 66.81 & 0.448 \\
\modelentry{qwen3-vl-4b} & 59.61 & 0.378 & 55.15 & 61.47 & 0.480 \\
\rowcolor{resultStripe}
\modelentry{qwen3-vl-2b} & 54.09 & 0.428 & 51.05 & 53.99 & 0.511 \\
\modelentry{deepseek-v4-flash-vision-exp} & 42.68 & 0.642 & 46.21 & 46.00 & 0.626 \\
\rowcolor{resultStripe}
\modelentry{glm-5-3-flash} & 74.99 & 0.311 & \underline{73.59} & \underline{82.45} & 0.423 \\
\modelentry{minicpm-v-4-6} & 48.01 & 0.517 & 44.23 & 51.53 & 0.544 \\
\rowcolor{resultStripe}
\modelentry{minicpm-v-4-5} & 37.59 & 0.583 & 32.01 & 39.06 & 0.552 \\
\modelentry{gemini-3-6-flash} & \textbf{75.73} & 0.310 & \textbf{76.27} & 81.96 & 0.406 \\
\rowcolor{resultStripe}
\modelentry{gemini-3-1-pro} & 71.98 & 0.300 & 68.62 & 77.26 & 0.386 \\
\modelentry{claude-opus-5} & \underline{75.27} & 0.294 & 71.24 & \textbf{84.01} & 0.400 \\
\rowcolor{resultStripe}
\modelentry{gpt-5-6-sol} & 61.59 & 0.472 & 62.26 & 69.75 & 0.502 \\
\modelentry{minimax-m3} & 42.54 & 0.628 & 43.10 & 47.29 & 0.605 \\
\rowcolor{resultStripe}
\modelentry{kimi-k2-6} & 68.02 & 0.335 & 62.44 & 75.14 & 0.481 \\
\modelentry{step3-vl} & 45.06 & 0.579 & 45.42 & 47.66 & 0.573 \\
\end{longtable}
}
\clearpage

\clearpage
\section{OmniDocBench Annotation Error Examples}
\label{app:annotation_errors}

\begin{figure}[!ht]
\centering
\includegraphics[width=\textwidth]{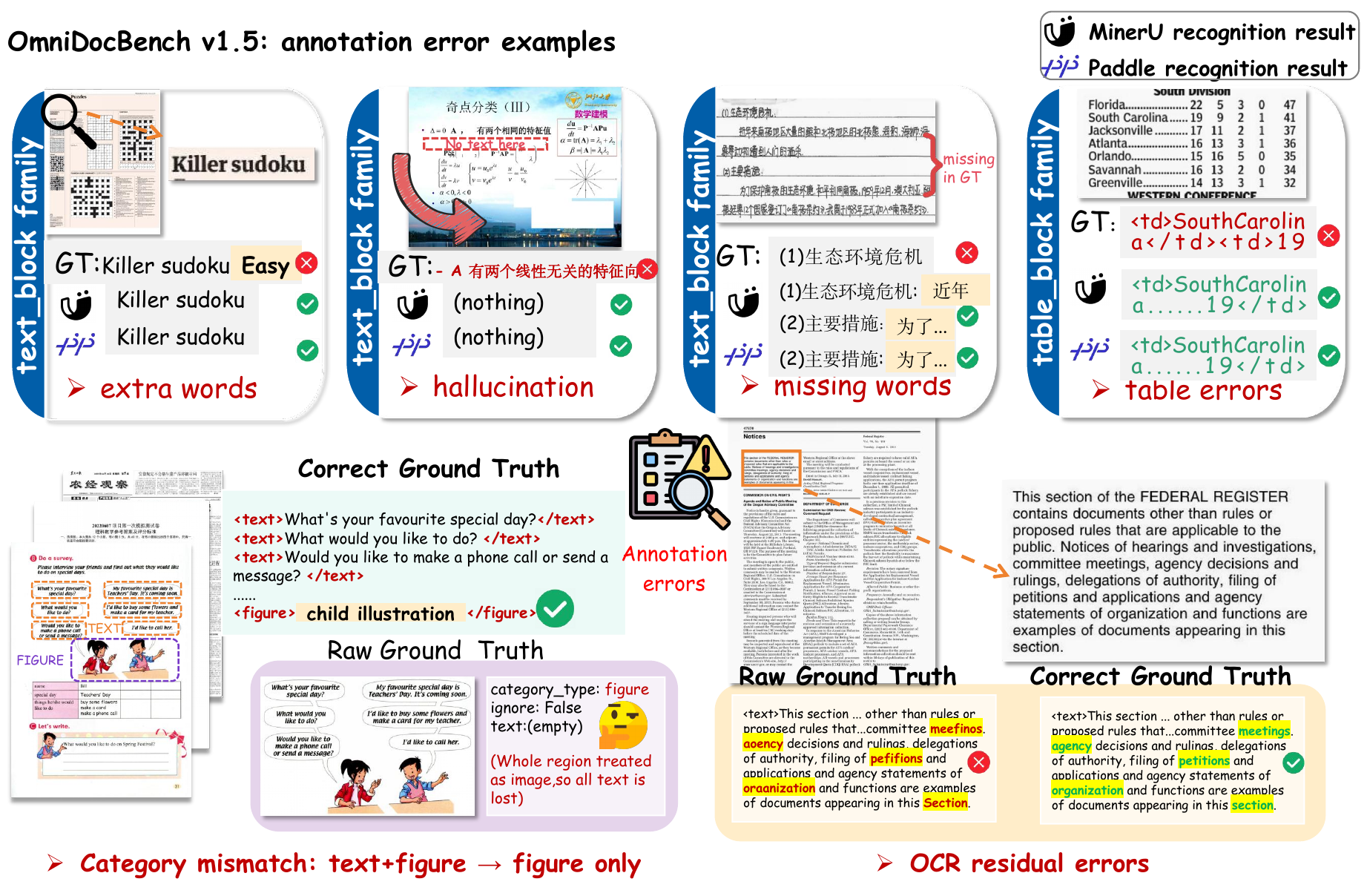}
\caption{\textbf{Annotation errors in OmniDocBench v1.5.} The examples cover extra words, hallucinated text, missing words, table structure errors, category mismatches, and residual OCR errors. Such discrepancies can affect the evaluation of otherwise correct predictions and change benchmark scores.}
\label{fig:audit_examples}
\end{figure}

Figure~\ref{fig:audit_examples} illustrates annotation discrepancies in OmniDocBench v1.5~\citep{omnidocbench}. Source-image regions are shown alongside annotation excerpts, covering extra words, hallucinated text, missing words, table structure errors, category mismatches, and OCR residual errors.

\clearpage
\section{Degradation Parameter Details}
\label{app:degradation_params}

\paragraph{Released degradation tracks.}
The released dataset keeps the three image tracks as parallel directory trees: clean pages under \texttt{images/}, digital degradation under \texttt{degraded\_v2\_merged/}, and physical recapture under \texttt{real\_degraded/}. All three tracks are GT-preserving: the file identity is mapped back to the same clean-page annotation, so differences in score isolate image-condition robustness rather than content changes. The digital generation manifest records \texttt{source=images}, \texttt{output=degraded\_v2}, \texttt{severity=moderate}, \texttt{severity\_val=0.6}, and zero generation failures. The ten digital scene families are first materialised separately under \texttt{degraded\_v2/<scene>/} with a per-image JSON parameter log under \texttt{degraded\_v2/params/<scene>/}; the evaluation track \texttt{degraded\_v2\_merged/} then selects exactly one scene version per clean page and records the provenance in \texttt{degraded\_v2\_merged/scene\_mapping.json}. This mapping is intentionally explicit: each evaluated digital image can be traced to its source clean page, scene name, generated PNG, and parameter JSON.

\paragraph{Digital scenes and operations.}
The digital track covers ten GT-preserving scene profiles spanning archival aging, bound-book scans, multi-generation copy chains, fax/thermal pages, low-quality printing, ink bleed, heavy compression, uneven lighting, geometric distortion, and noise--blur mixtures. These correspond to the release scene identifiers \texttt{aged\_archive}, \texttt{book\_binding}, \texttt{multi\_gen\_copy}, \texttt{fax\_thermal}, \texttt{low\_quality\_print}, \texttt{ink\_bleed}, \texttt{heavy\_compression}, \texttt{uneven\_lighting}, \texttt{geometric\_distort}, and \texttt{noise\_blur\_combo}. Each scene composes low-level OpenCV / NumPy operations such as paper grain, yellowing, foxing spots, faded ink, page curl, binding shadow, bleed-through, photocopier contrast, dirty-drum streaks, thermal/fax binarization, printer banding, ink dilation, JPEG compression, down-up sampling, lighting gradients, vignetting, perspective warp, lens distortion, rotation, Gaussian noise, motion blur, and warm/cool color shift. To avoid confounding content with image quality, no operation changes the underlying HTML source or GT ordering. The scene release is balanced by construction: round-robin assignment yields 147--148 pages per digital scene in the released \texttt{degraded\_v2} tree, and the merged track stores the selected scene for every clean page.

\begin{figure}[p!]
\centering
\includegraphics[width=\textwidth]{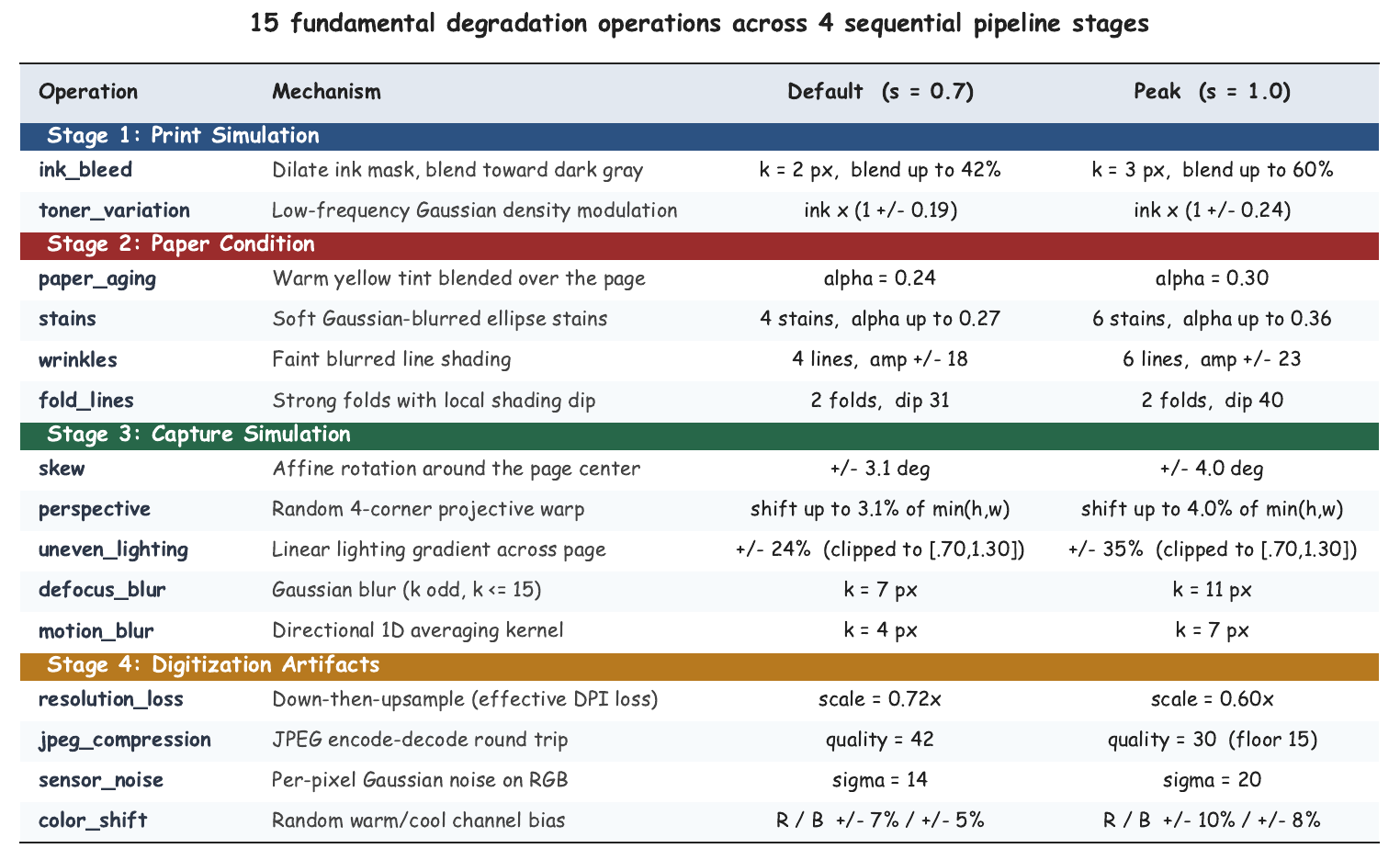}
\caption{\textbf{Fifteen degradation operations across printing, paper, capture, and digital processing.} The operations model changes in document appearance, with parameter values at ablation strength $s=0.7$ and peak stress $s=1.0$ supporting controlled tests of individual factors.}
\label{fig:degradation_ops}
\end{figure}

\begin{figure}[p!]
\centering
\includegraphics[width=\textwidth]{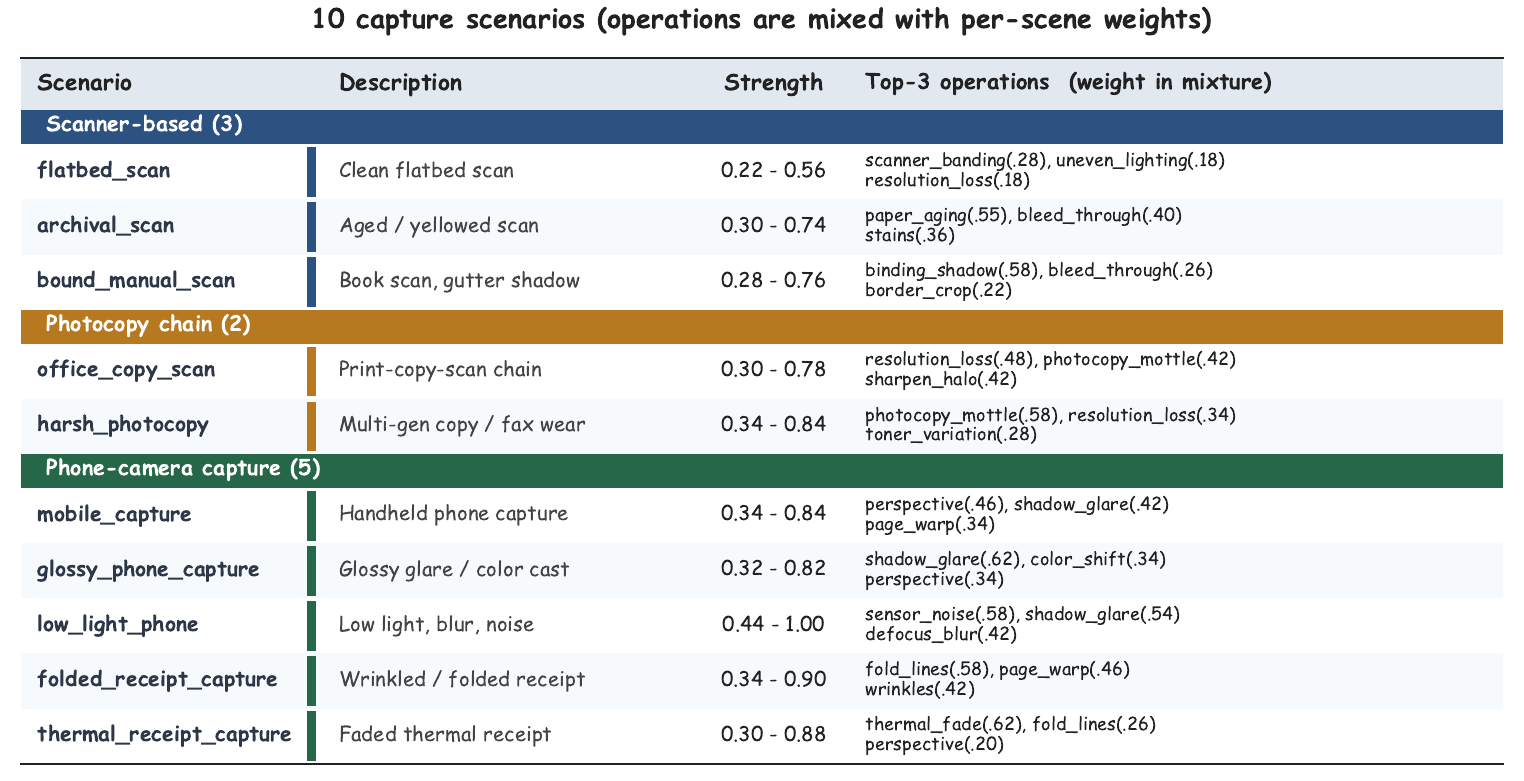}
\caption{\textbf{Ten capture scenarios combining multiple degradation factors.} Scenario-specific strength ranges and weighted operations represent different document acquisition conditions. The three dominant operations identify the main sources of degradation in each scenario.}
\label{fig:degradation_scenarios}
\end{figure}

\paragraph{Physical recapture and ablations.}
The real-degraded track is built from physical or screen-mediated recaptures of the rendered pages rather than synthetic filters. Filenames encode the acquisition condition after the \texttt{\_\_real\_} marker, e.g. phone/screen/screenshot source, paper state such as flat, creased, bent, cover, or photocopy, lighting such as normal, low-light, flash, shadow, outdoor, or screen, and view geometry such as top-down, oblique, handheld, or screenshot. The mapping files in the real-degraded evaluation workspace map each recaptured file back to the original clean relative path, again allowing the same GT to score all tracks. In addition to the released digital and real tracks, we keep an ablation generator that decomposes the degradation process into \textbf{15 fundamental operations} over a four-stage causal pipeline (\emph{Print $\to$ Paper $\to$ Capture $\to$ Digital}). Figure~\ref{fig:degradation_ops} lists these operations and their concrete parameter values at $s=0.7$ (single-factor ablation) and $s=1.0$ (peak stress). Figure~\ref{fig:degradation_scenarios} summarizes the higher-level capture scenarios and their dominant operation weights; these figures document the controllable design space used for sensitivity analysis, while \texttt{degraded\_v2\_merged/} is the canonical digital track used in the main benchmark.

\clearpage
\section{Full Per-Category Results}
\label{app:categories}

\ours{} pages are organized into ten top-level domains: \textit{academic}, \textit{education}, \textit{legal\_gov}, \textit{business}, \textit{finance}, \textit{medical}, \textit{publishing}, \textit{technical}, \textit{logistics}, and \textit{certificate}, with $180 / 138 / 151 / 160 / 212 / 200 / 100 / 110 / 120 / 104$ pages respectively (sum: $1{,}475$). For each (model, category, track) triple we recompute Overall as $((1{-}\mathrm{TextEdit})\times 100 + \mathrm{FormulaCDM} + \mathrm{TableTEDS})/3$ on the subset of samples in that category, and treat the entry as defined only when all three components exist (i.e.\ the category contains at least one text page, one formula sample, and one table sample). Per-category Avg$_3$ is then the mean of the three track Overalls.

Tables~\ref{tab:cat_academic}--\ref{tab:cat_certificate} report results for $53$ models across all ten domains. Each table includes Overall$\uparrow$, TextEdit$\downarrow$, FormulaCDM$\uparrow$, TableTEDS$\uparrow$, and ROEdit$\downarrow$ for Clean, Digital, and Real, followed by the three-track mean Avg$_3$. Models follow the architecture groups and within-series order of Table~\ref{tab:leaderboard}, with linked names, icons, and release months. Full-benchmark component scores appear in Appendix~\ref{app:full_results}. The category-level scores are provided in the supplementary file \texttt{category\_benchmark\_all\_tracks.csv}.

\paragraph{Cross-category observations.} Domain difficulty varies substantially across the $53$ models. Mean Avg$_3$ ranges from $42.0$ on \textit{certificate} to $82.9$ on \textit{academic}; individual scores span $9.4$--$68.9$ and $36.4$--$92.4$, respectively. Model dispersion is largest on \textit{certificate} ($\sigma{=}14.4$), followed by \textit{legal\_gov} ($13.5$), \textit{business} ($12.6$), and \textit{medical} ($12.4$). These differences motivate evaluating document parsers across domains with varied layouts and content.

\paragraph{Domain leaders and architecture mix.} GLM-5.3-Flash leads five domains by Avg$_3$: \textit{academic}, \textit{education}, \textit{legal\_gov}, \textit{business}, and \textit{medical}. Gemini 3.6 Flash leads \textit{publishing}; TeleOCR leads \textit{finance} and \textit{technical}; DotsMOCR leads \textit{logistics}; and FD-RL leads \textit{certificate}. Across the $50$ top-five positions ($10$ domains $\times$ $5$ ranks), general VLMs occupy $58\%$, E2E specialists $30\%$, and Pipeline specialists $12\%$. The domain winners comprise six VLM, three Pipeline, and one E2E result, showing that leadership depends on the document domain.

\clearpage
\input{data/per_category_tables.tex}

\end{document}